\documentclass[11pt]{article}

\usepackage[preprint]{acl}

\usepackage{times}
\usepackage{latexsym}

\usepackage{amsmath}
\usepackage{amssymb}
\usepackage[T1]{fontenc}

\usepackage[utf8]{inputenc}

\usepackage{microtype}

\usepackage{inconsolata}

\usepackage{graphicx}

\usepackage{listings}
\usepackage{xcolor}

\lstdefinestyle{promptstyle}{
    basicstyle=\ttfamily\small,
    breaklines=true,
    breakatwhitespace=true,
    frame=single,
    backgroundcolor=\color{gray!5},
    columns=fullflexible,
    keepspaces=true,
}
\definecolor{sftmint}{RGB}{232,247,236}
\definecolor{sftsand}{RGB}{248,244,230}
\definecolor{sftlavender}{RGB}{240,235,255}
\usepackage{float}
\usepackage{calc}

\usepackage{tabularx}   
\usepackage{siunitx}    
\usepackage{subcaption} 
\usepackage{listings}
\usepackage{booktabs}
\usepackage{xcolor}
\usepackage{colortbl}
\usepackage{array}
\usepackage{multirow}
\usepackage{enumitem}
\usepackage{tabularx}
\usepackage{pifont}
\usepackage{makecell}
\usepackage{graphicx}
\usepackage[table]{xcolor}

\usepackage[utf8]{inputenc}
\usepackage{longtable} 

\usepackage{booktabs}
\usepackage{multirow}
\usepackage{siunitx}
\usepackage{colortbl}
\usepackage{xcolor}
\definecolor{highlightcolorblue}{RGB}{225, 240, 255}
\definecolor{orange}{RGB}{255, 219, 187}
\usepackage{tabularx}
\usepackage{caption}  

\usepackage{siunitx}
\usepackage[table]{xcolor}

\usepackage{float}
\usepackage{siunitx}
\usepackage{twemojis}
\usepackage{fontawesome5}
\usepackage[dvipsnames]{xcolor} 
\usepackage{fontawesome5}       
\definecolor{highlightcolorblue}{HTML}{E6F0FF} 

\title{IntelliAsk: Learning to Ask High-Quality Research Questions via RLVR}

\author{Karun Sharma\thanks{Equal contribution.}, Vidushee Vats\footnotemark[1], Shengzhi Li\footnotemark[1], Yuxiang Wang, Zhongtian Sun, Prayag Tiwari}

\begin{document}
\maketitle
\begin{abstract}

Peer review relies on substantive, evidence-based questions, yet current LLMs generate surface-level queries that perform worse than human reviewer questions in expert evaluation. To address this gap, we curate a high-quality dataset of reviewer questions from OpenReview and conduct a human preference study where expert annotators evaluate question-paper pairs across three dimensions: effort, evidence, and grounding. From these annotations, we train IntelliReward, a reward model built from a frozen autoregressive LLM with trainable multi-head transformers. Validated against expert judgments, IntelliReward predicts reviewer-question quality better than API-based SFT baselines and provides scalable evaluation. We apply Decoupled Clip and Dynamic Sampling Policy Optimization (DAPO) with IntelliReward to train IntelliAsk, a question-generation model aligned with human standards of effortful, evidence-based critique. Human evaluations show IntelliAsk generates more grounded, substantive and effortful questions than strong baselines and reduces reliance on first-page content. We also find improvements on reasoning and writing benchmarks, suggesting reviewer-question quality correlates with broader capabilities. Compared to Qwen3-32B, IntelliAsk improves MuSR (68.3 vs 64.7 Acc) and WritingBench (8.31 vs 8.07). We release our code, filtered review dataset, expert annotations, IntelliAsk and IntelliReward to support automatic evaluation of grounding, effort, and evidence in LLM-generated review questions. (\url{https://anonymousse123456.github.io/intelliask.github.io/}
).
\end{abstract}

\section{Introduction}

Asking critical and well-reasoned questions is essential for advancing research, as such questions help clarify ideas, reveal limitations, and inspire new directions. In academic publishing, peer review plays a key role in this process, relying on reviewers to raise questions that improve the quality and impact of scientific work. However, as the number of submissions to major conferences has grown, the quality of reviewer feedback has declined. Many reviewers are overloaded and face tight deadlines, leading some to rely on large language models (LLMs) to draft questions and comments \citep{one}. While LLMs can produce fluent text, the questions they generate often lack technical depth, proper reasoning, or contextual understanding of the work.


\textbf{Why existing resources are not enough.} Most of the recent research works propose methods to improve the review generation capabilities of the LLMs. However, there's no focus on the quality of critic and the questions in the review generated by the models trained using these techniques,  hence rendering the review useless. Closer to our setting, \citet{opr}, fine-tunes LLaMA-8B on 79k reviews, but the generated questions extracted from the peer review just mimic the tone of reviewer style (See Section \ref{4}). The generated questions "sound" human, without offering a comprehensible and thoughtful question. \citet{autorev} uses a Graph based approach for generating peer reviews. While the graph structure helps organize paper content, the model still relies on simple supervised fine-tuning and produces questions that lack critical depth, remaining shallow imitations of human phrasing. Moreover, both \citet{opr} \& \cite{autorev} evaluate their systems primarily with automated review-quality scores from LLM judges, without incorporating human-in-the-loop assessments to measure whether the questions are actually useful to authors. Similarly, \citet{qasper} uses only titles and abstracts to generate questions, limiting the scope for creating technically detailed peer questions that are meaningful to authors. Overall, these approaches frame the task too broadly - treating it as generic review or QA generation-without explicitly modeling what makes reviewer questions effortful, evidence-based, and grounded.

\textbf{Challenges. }Generating effective review questions is not the same task as producing generic QA pairs based on the available content. LLM-generated questions often lack a clear understanding of technical content, resulting in questions that may be verbose and lengthy but unhelpful or already answered in the paper. Our own experiments highlight this gap:  we conducted an experiment where four expert annotators evaluated the questions generated by 3 strong baseline LLMs. They rated four variants of questions (3 model-generated and 1 human-written question from Openreview) each from o3, Gemini 2.5 Pro, Qwen2.5-32B and compared them to real human-authored questions. When evaluated with our rubric, \textbf{humans scored 0.78 points higher on average than the strongest model} and \textbf{1.53 points higher on average than the lowest scoring model} (see Table \ref{tab:human_eval}). The results show that human-written questions were consistently more relevant and useful. They were categorized to be written with more effort, contained evidence from the paper and weren't just framed using keywords from the paper, while the converse was true for the questions asked by the LLMs.

\textbf{Our Work.} In this paper, we address the challenge of generating critical, well-reasoned review questions.  We introduce an expert-annotated set of question-paper pairs scored on three metrics, and use it to train a reward model that serves as a scalable evaluation benchmark aligned with expert judgments. Finally, we show that while supervised fine-tuning (SFT) mostly imitates reviewer style, reinforcement learning guided by IntelliReward achieves closer alignment with human-authored questions.

Our contributions are as follows: 

\begin{enumerate}

    \item \textbf{Human Preference Data and IntelliReward}: We conduct a human annotation study with expert-annotated question--paper pairs evaluated across three criteria - Effort, Evidence, and Grounding. From this, we build IntelliReward, a reward model and automatic evaluation benchmark that aligns more closely with human judgment and outperforms API-based LLM-as-judge baselines tuned using SFT. To validate our reward model, we train 7B and 32B models using IntelliReward for quality critical question generation. 
    
    \item \textbf{IntelliAsk}: We develop a specialized question generation model trained using reinforcement learning (RL) to align with human standards. Unlike models trained with supervised fine-tuning (SFT) that primarily mimic stylistic tone, IntelliAsk asks technically deeper questions that significantly outperform SFT-only baselines and even exceed frontier models like Gemini 2.5 Pro in human evaluations. Furthermore, IntelliAsk demonstrates strong cross-task generalization, on external benchmarks for reasoning and general writing.

\end{enumerate}

\section{Question Extraction and Curation}
\label{2}

\subsection{Large-Scale Extraction of Questions from Openreview Reviews } 
\label{2.1}
We collected a dataset of reviewer feedback by scraping all publicly available reviews from ICLR 2024 using the OpenReview API. For each paper, we retrieved the corresponding metadata and downloaded the main PDF (excluding supplementary materials), limiting the maximum length to nine pages.\\
An Openreview submission includes several structured fields: \textit{Summary}, \textit{Strengths}, \textit{Weaknesses}, \textit{Questions}, \textit{Limitations}, \textit{Ethical Concerns}, numerical scores for \textit{Soundness} and \textit{Overall Evaluation}, and the reviewer's \textit{Confidence}. In practice, however, reviewers do not consistently confine their questions to the \textit{Questions} field.  
To characterize variability in question placement, we manually annotated a random sample of 100 reviews, observing that questions frequently appeared outside the designated \textit{Questions} section, sometimes they are present within the \textit{Weaknesses} or, less frequently, the \textit{Strengths} (See ~Fig\ref{fig:question_dist} in \ref{sec:question_placement}). In some cases, the \textit{Questions} section points to other sections (e.g., "See Weaknesses"), or mixed multiple questions with commentary. 

To address this variability and extract reviewer questions, we used Gemini 2.0 and prompted it with the concatenated text of the \textit{Questions}, \textit{Strengths}, and \textit{Weaknesses} sections from each review. The prompt explicitly instructed the model to copy questions verbatim, preserving their original phrasing and tone. (see \ref{extrac} in \ref{Appendix} for the full prompt). When a reviewer wrote multiple independent queries in a single sentence, the model split them into separate entries.  
To verify the accuracy, we manually inspected 500 extracted questions to ensure that the model consistently retained the original phrasing and did not hallucinate content.

After filtering, the final training dataset contained 15.5k questions drawn from 5,841 unique papers. The train dataset contains 13.2k questions and the test dataset contains 2.3k questions (see Appendix~\ref{sec:filtering_stages} for detailed filtering methodology and Figure~\ref{fig:waterfall_diagram} for the progressive filtering statistics).
To prepare the corresponding paper content for evaluation and training, we applied \texttt{olmOCR(olmOCR-7B-0825-FP8)}\citep{olmocr} to extract structured text from the first nine pages of each paper. 

\section{Benchmarking SOTA Reasoning LLMs Against Humans}
\label{3}
LLMs are capable of generating reviews when provided with a complete paper, however, they tend to fall short in asking compelling questions that involve critical thinking about the content of the paper and as well as the domain knowledge of the paper under consideration. To study this, we conduct a human annotation study comparing questions extracted from OpenReview reviews with those generated by several state-of-the-art LLMs.\\
\\
We primarily do this for below two reasons:
\begin{enumerate}
    \item To benchmark and quantify the gap between human and LLM-generated questions
    \item To create the preference data required to train a reward model for scaling annotation.
\end{enumerate}

\subsection{Human Preference and Annotation Study}
\label{3.1}
\textbf{Experimental Setup.}
Our preference study consists of 572 annotated question--paper pairs sampled from 300 randomly selected ICLR 2025 submissions on Openreview. For each paper, the full text was provided as input to the following large language models : Gemini 2.5 Flash (Reasoning model), o3 (Reasoning model), Qwen2.5-32B , under an identical prompting template (see \ref{qg}), yielding one model-generated question per system. In parallel, the corresponding human-authored reviewer question from Openreview was included as the reference.  
To eliminate source bias, all questions were anonymized before annotation. Human evaluators read each paper in full, including text, figures, and equations, to ensure proper context (See Fig \ref{fig:ui_screenshot} in Appendix for the User-Interface used by Annotators). If a paper was entirely outside an annotator's domain expertise, it was marked as skipped and reassigned. Annotators then scored each anonymized question according to the rubric introduced in Section~\ref{3.2}, which evaluates three binary dimensions: Effort, Evidence, and Grounding.

\subsection{Rubrics For Assessing Question Quality: Effort, Evidence, and Grounding}
\label{3.2}
To evaluate question quality, we design a rubric with three binary metrics: Effort, Evidence, and Grounding. Each metric is scored as 0/1, keeping the evaluation simple and consistent across annotators. We chose a binary scheme to reduce ambiguity and to focus on whether a question meets the essential qualities of being thoughtful and useful for authors. See \ref{sec:samples_rubric} for examples of each category.

\begin{enumerate}
    \item \textbf{Effort}: Does the question demand real thought to answer? 
    Low-effort questions can be answered by directly quoting the paper or restating surface-level details, whereas a high-effort question requires the reader to synthesize ideas, connect sections, or identify non-obvious implications beyond what is stated.

    \item \textbf{Evidence}: Is the question backed by specific content from the paper?  
    High-evidence questions point to particular results, assumptions, or arguments in the work and probe them critically.  
    Low-evidence questions raise points without support, making them speculative or unhelpful.

    \item \textbf{Grounding}: Is the question anchored in the actual content of the paper?  
    Grounded questions refer to concrete methods, experiments or claims across sections of the paper.  
    Ungrounded questions rely on generic phrasing, keywords or broad statements that could apply to almost any paper. For example: What if we increase the depth of the neural network ?
\end{enumerate}

\subsection{Analysis of Human vs. Model-Generated Questions}
\label{3.3}
\textbf{Source Vs Score.} 
Blind annotation results show that the Qwen2.5-32B model received the lowest scores, while highest quality human-authored questions from Openreview achieved the highest (see Table~\ref{tab:human_eval}). The mean cumulative score is calculated by taking an average of all the axis of the rubric, with the highest possible score being 3 and lowest 0. This gap becomes even clear when looking at the specific categories scores in Fig \ref{fig:fig3votedistribution}.

\textbf{First Page Bias (FPB).} 
We measure the fraction of words in the question that originate from the paper's first page. This tests whether models rely disproportionately on introductory text when framing questions. A high score indicates surface-level dependence, while lower scores suggest engagement with the full paper. Qwen2.5-32B shows the strongest dependence, with \textbf{55\%} of question words coming from the first page alone (Table~\ref{tab:human_eval}). In contrast, Human-authored questions, o3, and Gemini 2.5 Pro achieve relatively low scores, indicating that they draw more evenly from later sections of the paper when constructing questions. FPB is used only as an evaluation metric and is not part of the training objective.

\textbf{Question Length vs Source.}
Analysis of question length distributions across different sources reveals interesting patterns (see Figure~\ref{fig:kde_length_dist} in Appendix~\ref{sec:length_analysis}). Qwen2.5-32B produces the shortest questions, while Gemini 2.5 Pro generates the longest. The average length of o3's questions is close to that of Human-authored ones, but Humans show the highest variance, reflecting greater diversity and less reliance on fixed phrasing patterns.


\textbf{Question Length vs Score.}  
Comparing Human-authored questions with o3 reveals clear gaps in quality. For short questions (\(<20\) characters), Human-authored ones are more than \textbf{2$\times$ richer} in quality (effort + evidence + grounding) than those from o3. The largest gap is in grounding, where Humans outperform o3 by over \textbf{10$\times$}. Effort is also substantially lower for o3, suggesting that even its concise questions often lack depth and framing.

\section{SFT on Filtered Human Questions }
\label{4}

We fine-tuned \textit{Qwen/Qwen2.5-7B-Instruct-1M} on our curated training data using filtered Human-authored questions as the reference for reviewer-style generation. Training ran on four H200 GPUs for 24 hours with an input length of 14K tokens per paper. For evaluation, we held out a test split of 2200 samples from the curated data and used the same prompts as in our human annotation study to ensure fairness. During human evaluation, we observed that the SFT-generated questions were generally weak, we therefore additionally report automatic evaluation scores using our reward model, IntelliReward (see Section \ref{5.1} for IntelliReward).

The fine-tuned model learned to mimic the phrasing and tone of reviewers but did not improve in producing meaningful questions: depth, reasoning, and grounding remained weak compared to Human-authored questions (see Table \ref{tab:auto_eval}). We also tested existing SFT-trained reviewer models (OpenReviewer, DeepReviewer, AutoRev) by extracting the \textit{Questions} section of their outputs. Their results were fluent in style but shallow in substance, lacking the critical depth of Human-written questions (See \ref{examples}).  

These findings show that SFT captures style but not reasoning. High-quality reviewer questions require more than surface imitation, motivating our next step: RL with IntelliReward, a reward model trained to capture human preferences along Effort, Evidence and Grounding.

\section{Training IntelliAsk: A specialized model for asking critical questions}
\label{5}
As shown in Section \ref{4} and table \ref{tab:auto_eval}, SFT does not improve the model's performance on the critical question generation task. This limitation is consistent with recent findings showing that SFT often memorizes training data and struggles with out-of-distribution scenarios. Because of this tendency, it struggles to adapt to new situations. Reinforcement learning (RL), on the other hand, encourages exploration and learning from feedback, which helps it generalize better and handle tasks that require complex reasoning \citep{chu2025sft}.

\subsection{Reward Model : IntelliReward}
\label{5.1}

\begin{figure*}[t] 
    \centering
    \includegraphics[width=1\textwidth]{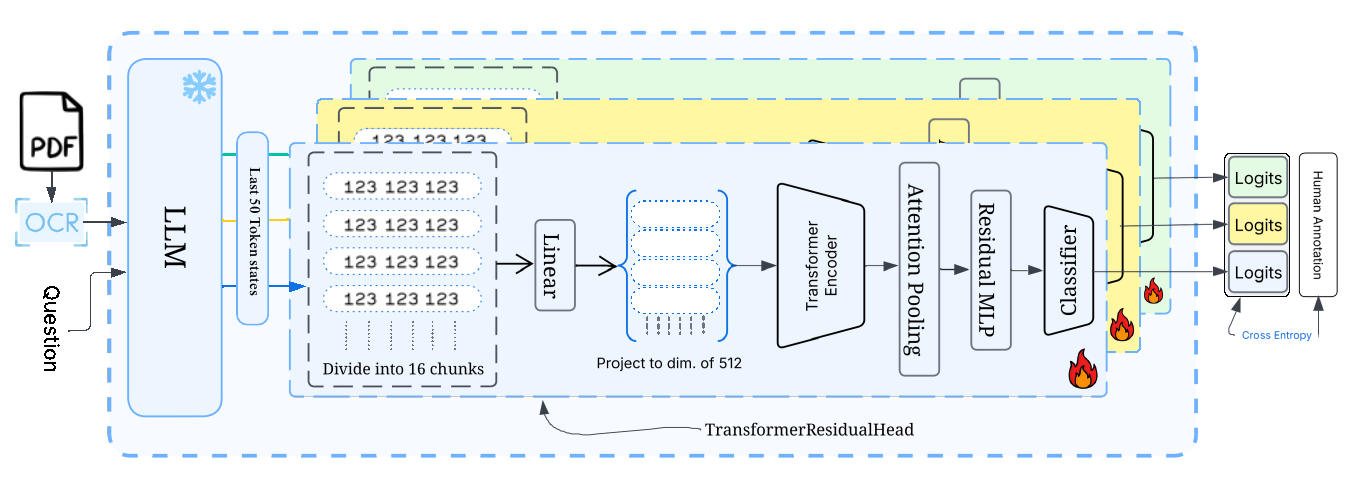}
    \caption{Architecture and training of the IntelliReward.}
    \label{fig:waterfall}
\end{figure*}

Evaluating all 15,500 questions with human annotators across three rubrics is costly and risks bias from fatigue. This highlights the need for a reliable automatic evaluation benchmark to support the scaling of our experiments. To reduce reliance on manual effort, we tested leading closed-source LLMs on the reward prediction task. However, they showed weak predictive accuracy (Table \ref{tab:extended_human_evaluation}), required large inputs, and incurred high inference costs, making them unsuitable for large-scale benchmarking. To overcome this, we trained IntelliReward on our human preference annotations to serve as an efficient and scalable substitute for human judgment. The architecture and training procedure are described in the following subsection.

\begin{table}[t]
\centering
\resizebox{\columnwidth}{!}{%
    \small
    \setlength{\tabcolsep}{3.5pt} 
    \renewcommand{\arraystretch}{1.2}
    \begin{tabular}{lccccc}
    \toprule
    & & \multicolumn{4}{c}{\textbf{Scores (\%)}} \\
    \cmidrule(lr){3-6} 
    
    \textbf{Model} & \textbf{Ckpt} & \textbf{Eff.} & \textbf{Evid.} & \textbf{Grd.} & \textbf{Acc.} \\
    \midrule
    
    \multicolumn{6}{l}{\textit{Closed-source LLMs (off-the-shelf)}} \\
    Gemini 2.5 Flash & Zero Shot & 57 & 25 & 29 & 37 \\
    GPT-4.1 & Zero Shot & 44 & 22 & 30 & 32 \\
    GPT-5 & Zero Shot & 56 & 54 & 49 & 53 \\
    \cmidrule(lr){1-6}
    
    \multicolumn{6}{l}{\textit{Closed-source LLMs (tuned with SFT)}} \\
    Gemini 2.5 Flash & SFT & 61 & 53 & 45 & 53 \\
    GPT-4.1 & SFT & 52 & 25 & 31 & 36 \\
    \cmidrule(lr){1-6}
    
    \multicolumn{6}{l}{\textit{Open-source baseline}} \\
    Qwen2.5-7B-Instr. & Original & 30 & 26 & 28 & 28 \\
    gpt-oss-20b & SFT & 44 & 32 & 35 & 37 \\

    \cmidrule(lr){1-6}
    
    \multicolumn{6}{l}{\textit{Our trained reward model}} \\
    \rowcolor{highlightcolorblue}
    \textbf{IntelliReward (ours)} & -- & \textbf{70} & \textbf{76} & \textbf{70} & \textbf{72} \\
    \bottomrule
    \end{tabular}%
}
\vspace{0.5em}
\caption{Reward prediction performance on the human preference annotation test split. We compare off-the-shelf models, SFT-tuned versions, and our IntelliReward. \textbf{Abbreviations:} Ckpt: Checkpoint, Eff.: Effort, Evid.: Evidence, Grd.: Grounding, Acc.: Acc is the average of per-dimension accuracies.}
\label{tab:extended_human_evaluation}
\end{table}

\subsection{Reward Model Architecture and Training}
\label{5.2}

\paragraph{Reward Model Architecture.}  

Our reward model handles multiple objectives by pairing a causal LLM with per-objective Transformer heads. We use \texttt{gpt-oss-20b} (medium reasoning) as the base. Given an input (e.g., paper OCR, generated question, task prompt), the LLM encodes it into a fixed representation. We extract the pooled hidden states of the last 50 output tokens and pass it to our per-objective Transformer head, which empirically improves performance as compared to using MLP head. (see Table \ref{tab:reward_arch_choice}). The resulting representation is denoted as
\[
r \in \mathbb{R}^{H}, \quad H = 2880,
\]
where \(r\) is the pooled hidden representation of the LLM outputs and \(H\) is its dimensionality.  

Each evaluation objective \(j \in \{1, \dots, k\}\) has an independent head \(f_j(\cdot)\) producing logits \(\ell_j \in \mathbb{R}^{C_j}\), where \(k\) is the total number of objectives and \(C_j\) is the number of classes (or possible labels) for objective \(j\). Each \texttt{TransformerResidualHead} first chunks \(r\) into \(n\) segments and projects them to dimension \(d_{\text{model}}\), then processes the sequence through \(L\) Transformer encoder layers. A learnable attention query pools the sequence into a vector \(z \in \mathbb{R}^{d_{\text{model}}}\), which is refined via a residual two-layer feedforward network (MLP):
\[
z' = \mathrm{LayerNorm}\!\bigl(z + \mathrm{FFN}(z)\bigr),
\]
where \(\mathrm{FFN}(\cdot)\) is the feedforward transformation and \(\mathrm{LayerNorm}(\cdot)\) denotes layer normalization. Finally, the refined vector is mapped to logits:
\[
\ell_j = W_j z' + b_j,
\]
where \(W_j \in \mathbb{R}^{C_j \times d_{\text{model}}}\) and \(b_j \in \mathbb{R}^{C_j}\) are learnable weights and biases for head \(j\).  

\paragraph{Training Objective and Inference}  
During training, the model minimizes the total loss $\mathcal{L} = \sum_{j=1}^{k} \mathrm{CE}(\ell_j, y_j)$, where $\mathrm{CE}$ denotes cross-entropy and $y_j$ is the ground-truth label for objective $j$. During inference, each head predicts $\hat{y}_j = \arg\max \ell_j$, and the final score is computed as $S = \sum_{j=1}^{k} \hat{y}_j$.

\paragraph{Reward Model Training.}
We train IntelliReward using the human preference annotations collected in our study. 
The frozen LLM provides representations, while only the per-objective heads $f_j(\cdot)$ are updated. 
Training follows the cross-entropy loss defined above. 
We optimize with AdamW (learning rate $2 \times 10^{-5}$, batch size $8$, weight decay $0.01$) for $5$ epochs on a single NVIDIA L40S GPU. 
End-to-end training completes within $30$ minutes. 
The Per-objective Head is  lightweight and only takes total of 300MB of GPU VRAM during inference.

\newcolumntype{Y}{>{\raggedright\arraybackslash\hsize=1.3\hsize}X}



\begin{table}[t]
\centering
\small
\setlength{\tabcolsep}{3.5pt} 
\renewcommand{\arraystretch}{1.1} 
\setlength{\aboverulesep}{1.5pt}
\setlength{\belowrulesep}{1.5pt}

\begin{tabular}{clcccc}
\toprule
& & \multicolumn{4}{c}{\textbf{Scores (\%)}} \\
\cmidrule(lr){3-6}
\textbf{Base} & \textbf{Pool} & \textbf{Eff.} & \textbf{Evid.} & \textbf{Grd.} & \textbf{Mean} \\
\midrule

\multicolumn{6}{l}{\textit{Head: Standard MLP}} \\
\twemoji{snowflake} & None & 61 & 64 & 61 & 62 \\
\twemoji{snowflake} & Pool50 & 64 & 67 & 64 & 65 \\
\twemoji{fire} & None & 64 & 65 & 60 & 63 \\
\twemoji{fire} & Pool50 & 65 & 69 & 67 & 67 \\
\twemoji{fire} & Pool128 & 64 & 68 & 66 & 66 \\
\midrule 

\multicolumn{6}{l}{\textit{Head: Transformer Residual (Ours)}} \\
\twemoji{snowflake} & None & 68 & 68 & 70 & 69 \\
\twemoji{snowflake} & Pool50 & 70 & 76 & 70 & 72 \\
\twemoji{snowflake} & Pool128 & 69 & 77 & 67 & 71 \\
\twemoji{fire} & None & 71 & 69 & 70 & 70 \\
\rowcolor{highlightcolorblue}
\twemoji{fire} & Pool50 & \textbf{71} & \textbf{78} & \textbf{70} & \textbf{73} \\
\twemoji{fire} & Pool128 & 70 & 78 & 68 & 72 \\
\bottomrule
\end{tabular}
\vspace{0.2em} 
\caption{Ablation study comparing head architectures. \textbf{Base:} \twemoji{snowflake}\,= Frozen backbone, \twemoji{fire}\,= Trainable backbone. \textbf{Pool:} Pooling strategy ($k$=last $k$ tokens). \textbf{Scores:} Eff.=Effort, Evid.=Evidence, Grd.=Grounding.}
\label{tab:reward_arch_choice}
\end{table}



\subsection{RL using IntelliReward Reward Model}
\label{5.3}

As shown in Section~\ref{4}, supervised fine-tuning (SFT) performs poorly for review question generation: the model copies surface style but does not produce questions with real effort, evidence, or grounding. To address this, we use our reward model, \textbf{IntelliReward}, to align generation with human preferences. Fig \ref{fig:rew} shows the difference in reward curve for both Qwen2.5-7B-1M and IntelliAsk. \\

We train IntelliAsk-7B with  DAPO\citep{dapo} and IntelliAsk-32B with GRPO. For each paper, the model generates several candidate questions, which are scored by IntelliReward, and these scores are used as rewards to guide optimization. Training follows the standard DAPO and GRPO setup (batch sizes, sequence length, gradient clipping, learning rate schedule; see Appendix~\ref{sec:hyperparameters}). The resulting model, \textbf{IntelliAsk-32B}, consistently outperforms SFT-only baselines by producing questions that are more evidence-based, better grounded, and require greater effort.

\section{Evaluation}
\label{sec:evaluation}

We evaluate IntelliAsk across three key dimensions: (1) \textbf{Human Evaluation} to measure quality through expert assessment on the three rubric dimensions, and (2) \textbf{Automatic Evaluation} using IntelliReward to scale evaluation across larger test sets and external benchmarks (3)  Generalization to broader writing tasks beyond scientific question generation.
Each rubric (Effort, Evidence, Grounding) is labeled as a binary variable, reported values are means across samples, and Total is the sum of the three means. First Page Bias (FPB) is the fraction of content words in the question that overlap with the OCR text from page 1 (lowercased, stopwords removed) as defined in \ref{3.3}
\subsection{Human Evaluation}
\label{sec:human_eval}

To validate the quality of generated questions, we conducted a blind human evaluation study on more than 100 randomly sampled papers from the test set. Four expert annotators evaluated questions from multiple systems according to our three-dimensional rubric (Effort, Evidence, Grounding). 

Table~\ref{tab:human_eval} presents the human evaluation results. Human-authored questions from OpenReview achieve the highest scores across all dimensions, with a total score of 1.57/3.0, demonstrating substantial effort, evidence-based reasoning, and grounding in paper content. Among models, our IntelliAsk-32B achieves a score of 0.66/3.0, outperforming Gemini 2.5 Pro (0.60). Notably, IntelliAsk-32B achieves the lowest first page bias (21.37\%), indicating that it draws from the full paper rather than relying primarily on the introduction. Baseline models like Qwen2.5-32B perform poorly (0.05/3.0), confirming that standard pretraining without targeted alignment fails to produce thoughtful reviewer questions.

\begin{table}[t]
\centering
\footnotesize
\setlength{\tabcolsep}{3.5pt}
\renewcommand{\arraystretch}{1.1}
\begin{tabular}{
  l
  c
  S[table-format=1.2]
  S[table-format=1.2]
  S[table-format=1.2]
  S[table-format=1.2]
  S[table-format=2.2]
}
\toprule
\textbf{Model} &
\textbf{Rsn.} &
\multicolumn{3}{c}{\textbf{Scores [0--1]}} &
\textbf{Total} &
\textbf{FPB.} \\
\cmidrule(lr){3-5}
 & & \textbf{Eff.} & \textbf{Evid.} & \textbf{Grd.} & & \textbf{(\%) $\downarrow$} \\
\midrule

\multicolumn{7}{l}{\textit{\textbf{Human-Evaluated Scores}}} \\
\addlinespace[2pt]

\rowcolor{gray!10}
Human questions & -- & 0.54 & 0.46 & 0.57 & 1.57 & 28.21 \\
o3 & Med. & 0.32 & 0.12 & 0.36 & 0.80 & 16.81 \\
Gemini~2.5~Pro & Def. & 0.26 & 0.13 & 0.21 & 0.60 & 25.75 \\
IntelliAsk-32B & Def. & 0.27 & 0.13 & 0.26 & 0.66 & 21.37 \\
Qwen2.5-32B & No & 0.02 & 0.01 & 0.02 & 0.05 & 54.96 \\

\bottomrule
\end{tabular}

\vspace{0.3em}
\caption{
Human evaluation on ICLR~2024 papers.
Scores are Effort (Eff.), Evidence (Evid.), and Grounding (Grd.), each in $[0,1]$.
Reasoning modes: Medium (Med.), Default (Def.).
First Page Bias (FPB.)lower is better.
}
\label{tab:human_eval}
\end{table}

\subsection{Automatic Evaluation with IntelliReward}
\label{sec:auto_eval}

\begin{table*}[t]
\centering
\small
\setlength{\tabcolsep}{7pt}
\renewcommand{\arraystretch}{1.05}
\begin{tabular}{
  l 
  c 
  S[table-format=1.2] 
  S[table-format=1.2] 
  S[table-format=1.2] 
  S[table-format=1.2] 
  S[table-format=2.2] 
}
\toprule
{\multirow{2}{*}{\textbf{Model / Source}}} &
{\multirow{2}{*}{\textbf{Reasoning}}} &
\multicolumn{3}{c}{\textbf{Scores [0--1]}} & 
{\multirow{2}{*}{\textbf{Total [0--3]}}} & 
{\textbf{FPB.}} \\
\cmidrule(lr){3-5}
 & & {\textbf{Effort}} & {\textbf{Evidence}} & {\textbf{Grounding}} & & {\textbf{(\%) $\downarrow$}} \\ 
\midrule

\multicolumn{7}{l}{\textit{\textbf{Large Models}}} \\
\addlinespace[2pt]
gpt-oss-120b & Medium & 0.08 & 0.15 & 0.12 & 0.35 & 22.99 \\
gpt-4.1 & No & 0.07 & 0.12 & 0.12 & 0.31 & 31.73 \\
gpt-5 & Default & 0.09 & \textbf{0.20} & 0.16 & 0.45 & 18.63 \\
\rowcolor{gray!10}
o3 & Medium & \textbf{0.28} & 0.14 & \textbf{0.30} & \textbf{0.72} & 16.81 \\
claude-3.7-sonnet & No & 0.09 & \underline{0.18} & 0.15 & 0.42 & 45.14 \\
claude-3.7-sonnet & Default & 0.08 & 0.16 & 0.13 & 0.37 & 47.13 \\
gemini-2.5-flash & No & 0.08 & 0.15 & 0.15 & 0.38 & 39.06 \\
gemini-2.5-pro & Default & 0.22 & 0.11 & 0.18 & 0.51 & 25.75 \\
llama-4-maverick & No & 0.09 & 0.17 & 0.15 & 0.41 & 48.48 \\
grok-4 & No & 0.07 & 0.14 & 0.12 & 0.33 & 35.47 \\
deepseek-chat-v3.1 & Default & 0.11 & \textbf{0.20} & 0.17 & 0.48 & 36.83 \\

\addlinespace[8pt]

\multicolumn{7}{l}{\textit{\textbf{Small Open-Source Models ($\le$ 32B)}}} \\
\addlinespace[2pt]
\rowcolor{sftsand}
OpenReviewer-8B & No & 0.00 & 0.00 & 0.10 & 0.10 & 51.14 \\
\rowcolor{sftsand}
DeepReviewer-7B & No & 0.00 & 0.00 & 0.10 & 0.10 & 48.14 \\
gpt-oss-20b & Medium & 0.06 & 0.11 & 0.10 & 0.27 & 24.81 \\
Qwen2.5-7B & No & 0.00 & 0.01 & 0.01 & 0.02 & 49.93 \\
\rowcolor{sftsand}
Qwen2.5-7B SFT (Ours) & No & 0.00 & 0.01 & 0.02 & 0.03 & 42.11 \\
IntelliAsk-7B (Ours) & No & 0.03 & 0.07 & 0.07 & 0.17 & 27.44 \\

Qwen3-32B & Default & 0.05 & 0.13 & 0.09 & 0.28 & 26.73 \\
\rowcolor{highlightcolorblue}
IntelliAsk-32B (Ours) & Default & \underline{0.23} & 0.12 & \underline{0.20} & \underline{0.55} & 21.37 \\

\bottomrule
\end{tabular}
\vspace{0.3em}
\caption{Automatic evaluation using IntelliReward on test set. Rows highlighted in beige correspond to SFT baseline models (OpenReviewer-8B, DeepReviewer-7B, and Qwen2.5-7B SFT). IntelliAsk-32B achieves the highest score among small models (0.55/3.0), substantially outperforming SFT-only baselines. Among all models, o3 achieves the best performance.
\textbf{Bold}: best in category; \underline{underline}: second-best. FPB = First Page Bias}
\label{tab:auto_eval}
\end{table*}

\subsection{Generalization to Writing Tasks}
\label{sec:generalization}

Beyond scientific question generation, we evaluate whether the skills learned by IntelliAsk transfer to general writing and reasoning tasks. Table~\ref{tab:external_benchmarks} presents results across multiple benchmarks spanning reasoning, comprehension, and writing domains against its base model.

\textbf{Reasoning \& Comprehension:} IntelliAsk-32B achieves strong performance on reading comprehension \citep{drop,boolq} and multi-step reasoning \citep{musr,gpqa}, matching or exceeding the baseline Qwen3-32B model. This suggests that learning to ask evidence-based questions enhances the model's ability to understand and reason about complex content.

\textbf{Writing \& Generation:} Most notably, IntelliAsk-32B outperforms Qwen3-32B on WritingBench \citep{writingbench} and Arena Hard \citep{arenahard}, demonstrating that training on high-quality question generation improves general writing ability. This supports our core thesis: learning to ask better questions transfers to better writing across diverse domains.

\begin{table}[t]
\centering
\footnotesize
\setlength{\tabcolsep}{4pt}
\renewcommand{\arraystretch}{1.1}
\begin{tabular}{lccc}
\toprule
\textbf{Benchmark} & \textbf{IA-32B} & \textbf{Qwen3-32B} & \textbf{Metric} \\
\midrule

\rowcolor{gray!10}
\multicolumn{4}{l}{\textit{\textbf{Reasoning \& Comprehension}}} \\
DROP & \textbf{95.1} & 93.3 & F1 / Acc \\
MuSR & \textbf{68.3} & 64.7 & Acc \\
BoolQ & \textbf{90.0} & \textbf{90.0} & Acc \\
GPQA-Diamond & \textbf{69.1} & 68.4 & Acc \\

\addlinespace[3pt]
\rowcolor{gray!10}
\multicolumn{4}{l}{\textit{\textbf{Writing \& Generation}}} \\
WritingBench & \textbf{8.31} & 8.07 & 0--10 \\
Arena Hard & \textbf{94.1} & 93.8 & 0--100 \\

\bottomrule
\end{tabular}

\vspace{0.3em}
\caption{
Generalization on external benchmarks.
IA-32B (IntelliAsk-32B) outperforms Qwen3-32B on writing tasks (WritingBench, Arena Hard) while remaining competitive on reasoning and comprehension benchmarks. Learning to ask better questions improves general writing ability.
}
\label{tab:external_benchmarks}
\end{table}

These results demonstrate that IntelliAsk not only excels at scientific question generation but also improves general language understanding and writing capabilities.

\section{Related Work}
\looseness=-1
Recent research has increasingly explored the use of large language models (LLMs) to automate aspects of peer review. Several works train models on large corpora of reviews, often through supervised fine-tuning (SFT). For instance, \citet{opr} introduce \textit{OpenReviewer}, fine-tuning LLaMA-8B on 79K reviews to produce fluent and structured assessments, while \citet{zhu-etal-2025-deepreview} develop \textit{DeepReview}, a multi-stage pipeline that integrates retrieval and self-reflection, supported by the curated DeepReview-13K dataset. Similarly, \citet{tan2025peer} propose \textit{ReviewMT}, a dataset of 110K review comments enabling multi-turn, role-based review dialogue. While these systems improve stylistic fluency and tone, they primarily focus on generating full reviews rather than isolating and producing the probing questions or issue-driven feedback that most benefits authors.

\looseness=-1
Other approaches explore multi-agent frameworks. \citet{darcy2024margmultiagentreviewgeneration} propose \textit{MARG}, which distributes paper sections across specialized agents (e.g., clarity, experiments, impact) that collaborate to generate comprehensive feedback, mitigating context-length limitations. Similarly, \citet{chamoun-etal-2024-automated} introduce \textit{SWIF$^2$T}, which decomposes review generation into planner, investigator, reviewer, and controller modules to provide focused, actionable comments. These approaches enhance specificity and helpfulness relative to earlier baselines that mostly generate general feedback or superficial style corrections.

\looseness=-1
Several datasets and evaluation frameworks also relate closely. \citet{baumgartner-etal-2025-peerqa}, \citet{sundar2024cpapers}, and \citet{singh-etal-2024-scidqa} harvest reviewer questions and author responses-facilitating tasks such as answer generation or content retrieval rather than explicit question generation itself. On the evaluation side, recent work such as GEM PiCO \citep{ning2025pico}, and ReviewCritique \citep{du-etal-2024-llms} analyze the quality of reviews via off-the-shelf LLM judges or annotated corpora, focusing on fluency and consistency. Almost all of these works rely on SFT or prompting, and none explicitly train a model purely for reviewer-style question generation using human-labeled question data.

\looseness=-1
Despite this progress, existing research overwhelmingly treats peer review as a problem of generating full reviews or answering reviewer questions. Very little attention has been given to \textit{question generation itself}---the actionable and constructive element of peer feedback. Moreover, the dominant reliance on SFT or LLM-as-judge evaluations leaves a gap in aligning generation with the qualities that authors value most: effortful engagement, grounded critique, and context-aware probing. Our work directly addresses this gap by introducing a human-annotated dataset of reviewer-style questions, and by training with supervised fine-tuning to generate them, thereby offering a new benchmark and model geared specifically toward generating probing, useful questions in peer review.

\section{Conclusion}
\looseness=-1
We show that generating high-quality reviewer questions is a distinct and challenging capability that is not captured by supervised fine-tuning alone. Through expert annotations, we formalize question quality along three dimensions: effort, evidence, and grounding, and use them to train IntelliReward, a scalable reward model that aligns closely with human judgment and significantly outperforms API based LLM-as-judge. Using this, we train IntelliAsk using reinforcement learning and demonstrate substantial gains over SFT-based and frontier baselines in both human and automatic evaluations. Beyond peer review, IntelliAsk shows consistent improvements on external reasoning, comprehension, and writing benchmarks, indicating that learning to ask better questions transfers to broader language abilities. These results suggest that high-quality questions serves as a meaningful proxy for deeper understanding and reasoning. 


\section*{Limitations}
\looseness=-1
A natural extension of this work is to include multimodal content like figures and diagrams, and to evaluate the approach across more research domains and conferences. Further scaling IntelliAsk to larger foundation models and more importantly more human annotation will greatly improve the models capabilities for asking high quality research questions. We just have to be careful that reviewers don't use the most complicated questions generated by our LLMs as an excuse to fail a paper.


\section*{Ethical Consideration and Data Licensing}
\looseness=-1
The dataset was created from publicly available reviewer comments on ICLR papers hosted on Openreview.net. We restricted the collection to publicly accessible text and removed any metadata that could identify reviewers. As OpenReview content is distributed under the CC BY 4.0 license, our use and release of these comments complies with the original terms.
The human preference annotations are original contributions and are released under the same CC BY 4.0 license. We do not claim copyright over the original review texts or paper excerpts.

\bibliography{custom}

\appendix

\section{Appendix}
\label{Appendix}

\subsection{Multi-Stage Filtering Process}
\label{sec:filtering_stages}

\begin{figure*}[h]
  \centering
  \includegraphics[width=0.5\linewidth]{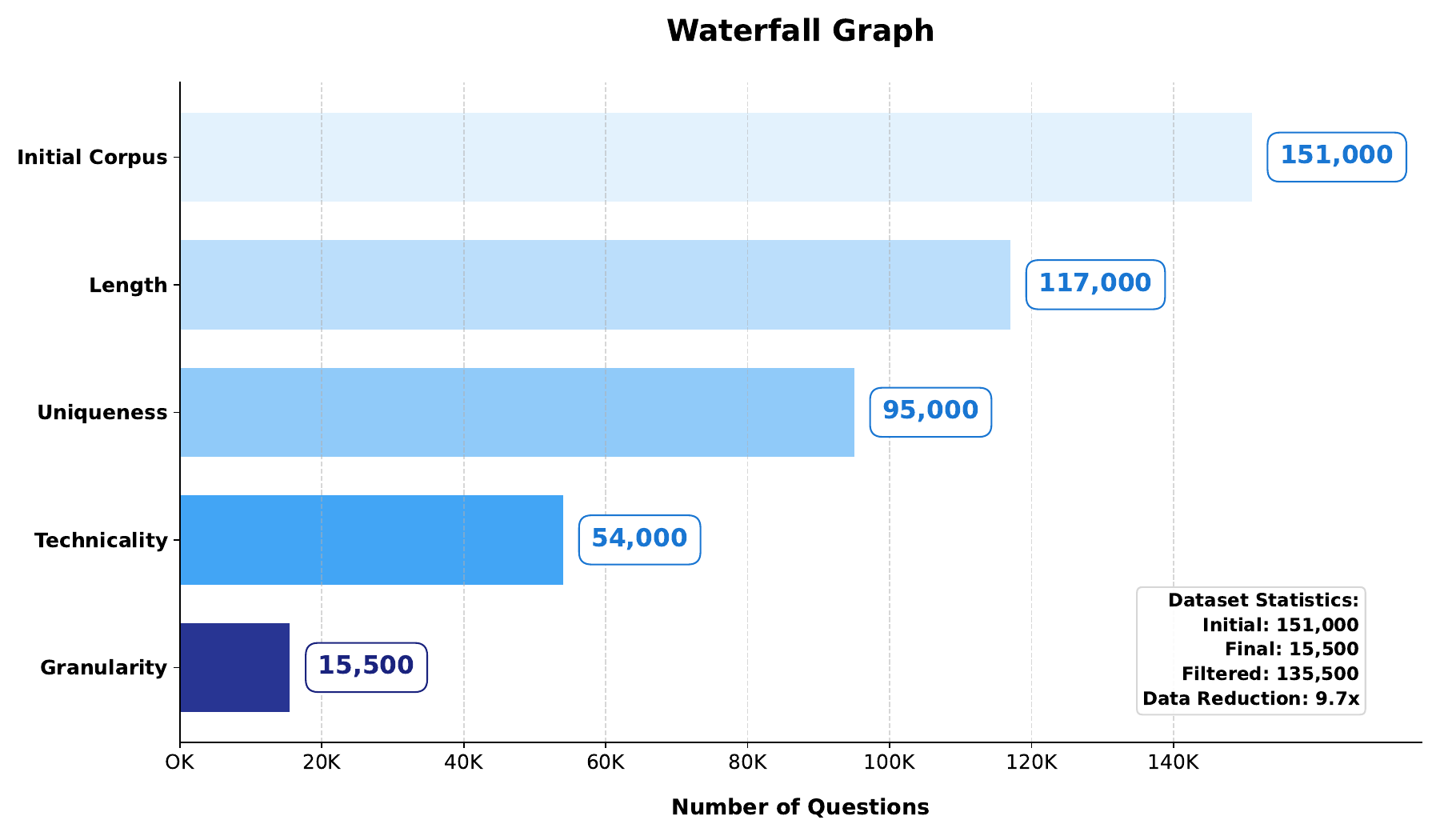}
  \caption{Waterfall diagram illustrating progressive instance filtering at each stage of the data curation process.}
  \label{fig:waterfall_diagram}
\end{figure*}

To have a dataset suitable for downstream modeling, we applied a series of filtering steps guided by best practices from CVPR reviewer slides \citep{cvpr_slides}, NeurIPS \citep{neurips_review} and ICLR reviewer guidelines \citep{iclr_review}, prior work on LLM feedback for reviews \citep{paper}, and our own manual inspection of roughly 2,000 reviews. The initial extraction produced about 151,000 questions. Our goal was not simply to maximize quantity but to ensure that the retained questions were clear, specific, and technically relevant. Each filtering stage systematically removed low-quality or redundant entries. After every stage, we manually checked a random sample of about 1,000 questions to confirm that the filtering criteria were effective and that valid questions were not being discarded.

\textbf{Length-Based Filtering.}
We first excluded questions under 100 characters. Manual analysis showed that short questions typically contained superficial comments or clarifications readily apparent in the submission text. This filtering step removed 34,000 entries, resulting in a subset of 117,000 questions. We then proceed to remove semantically similar questions.

\textbf{Eliminating Semantically Redundant Questions.}
Numerous questions were semantically identical apart from minor variations in wording. Training on highly redundant content increases the risk of overfitting and limits output diversity. To address this, we applied clustering using Stella with a cluster size of k=5. This reduced the dataset to 95,000 questions. After this stage of filtering there were still many questions which were non-technical and not relevant to the content of the paper for which we employ another stage of filtering described further.

\textbf{Filtering Non-Technical and Irrelevant Content.}
Manual review identified many questions unrelated to the technical content, including remarks on grammar, formatting, typographic errors, and unprofessional or subjective comments. Prior work \citep{one} has shown that reviews containing certain keywords (e.g., "commendable," "innovative") are often generated by language models. To mitigate this, we developed a prompt specifying six exclusion criteria, detailed in the Appendix(See \ref{qg3}). Importantly, we provided Gemini 2.0 Flash with both the review text and the corresponding paper as context, ensuring that ungrounded or off-topic questions could be more reliably detected and filtered. This process removed 41,000 questions. Even after this stage, we observed remaining questions that were purely opinion-based or that dismissed techniques without justification, which were addressed in the subsequent filtering stage.

 \textbf{Filtering for Specificity and Actionability.}
The final stage removed questions that were vague or speculative. We targeted two categories:
(i) incomplete, rhetorical, or opinion-based questions without supporting evidence;
(ii) unsupported assertions that a technique would fail or had been previously published (See \ref{qg4} in \ref{Appendix}). Questions were sequentially evaluated, retaining only those that satisfied all criteria. This step removed 38,500 questions, resulting in a final corpus of approximately 15,500 diverse, technically relevant entries.

\subsection{SFT vs RL Training Curve}

\begin{figure}[H] 
  \centering
  \includegraphics[width=\columnwidth]{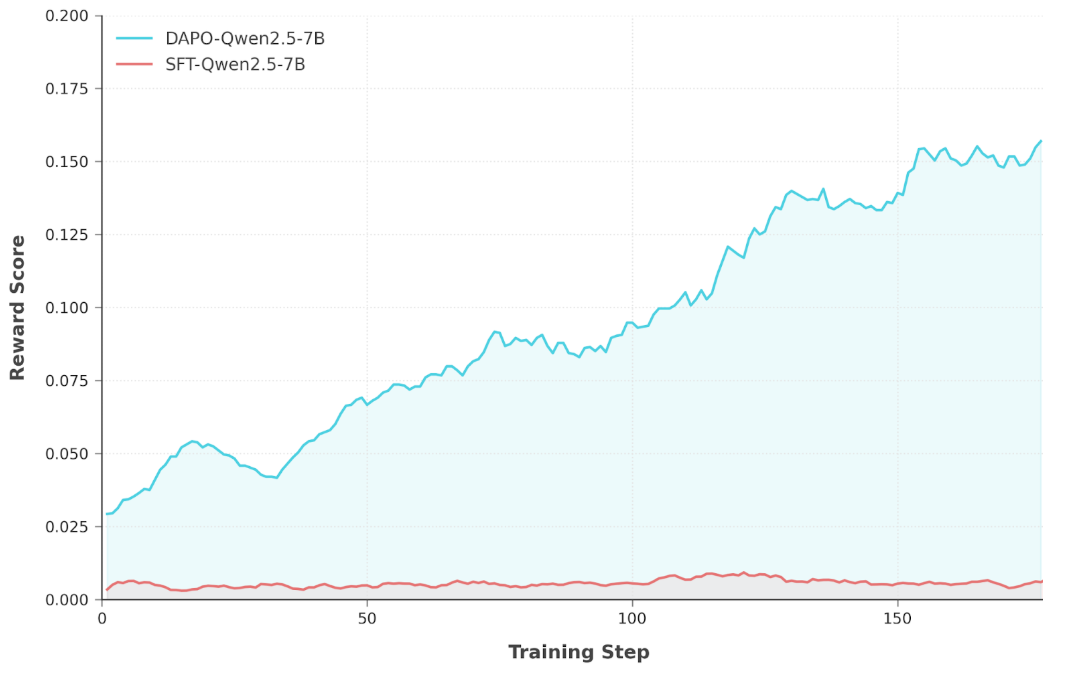}
  \caption{The figure show the difference in reward curves for Qwen2.5-7B (SFT) and IntelliAsk during training.}
  \label{fig:rew}
\end{figure}

\subsection{Question Length Distribution Analysis}
\begin{figure}[h]
  \centering
  \includegraphics[width=\columnwidth]{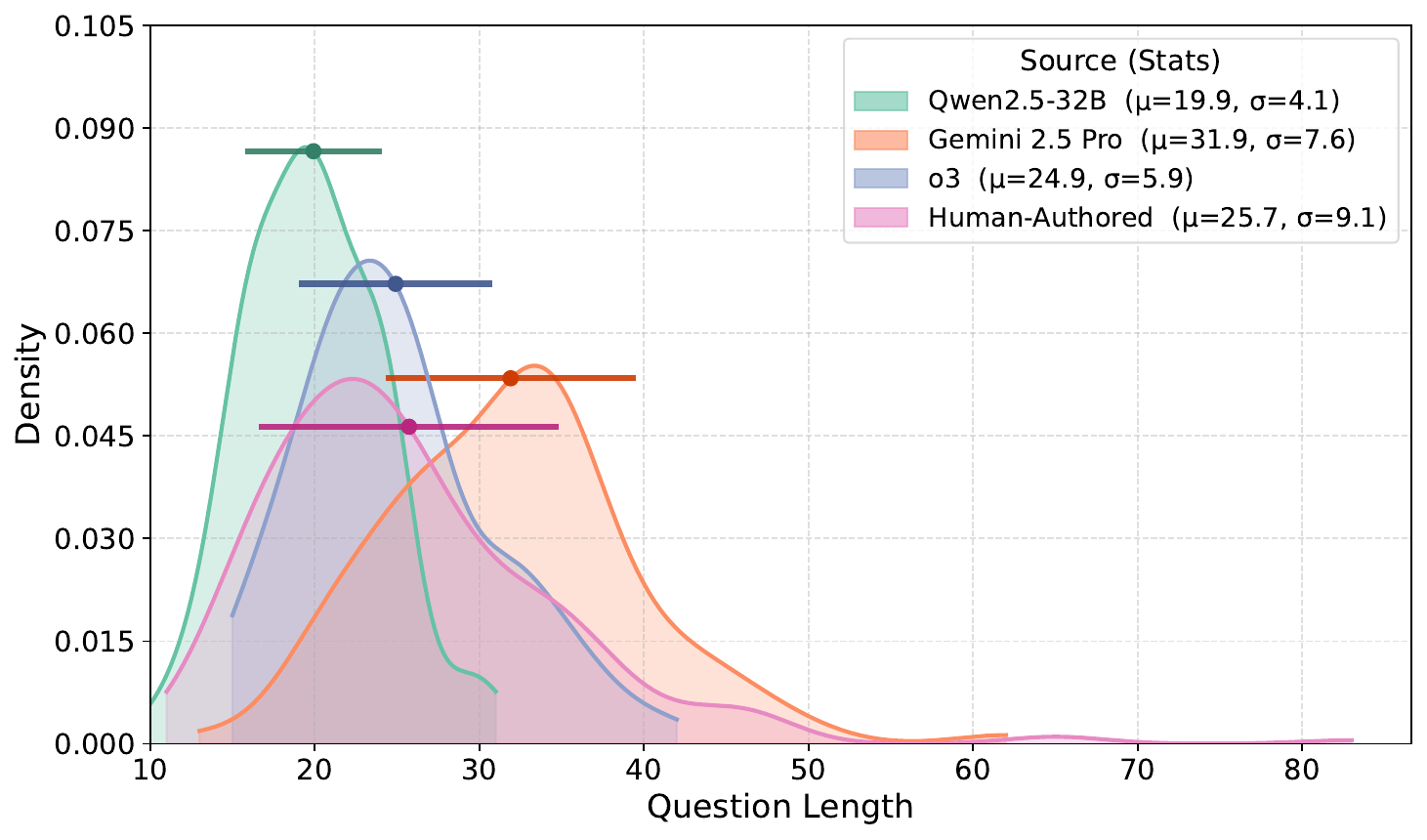}
  \caption{Distribution of question lengths across sources. Kernel density estimates show that human-authored questions exhibit the highest variance, reflecting greater diversity. Qwen2.5-32B produces the shortest questions, while Gemini 2.5 Pro generates the longest.}
  \label{fig:kde_length_dist}
\end{figure}

Fig \ref{fig:kde_length_dist} shows the distribution of the length of questions generated by the models against the questions written by Reviewers.
\label{sec:length_analysis}

\subsection{Distribution of votes on Effort, Evidence and Factual metrics by source}

\begin{figure}[h]
  \includegraphics[width=\columnwidth]{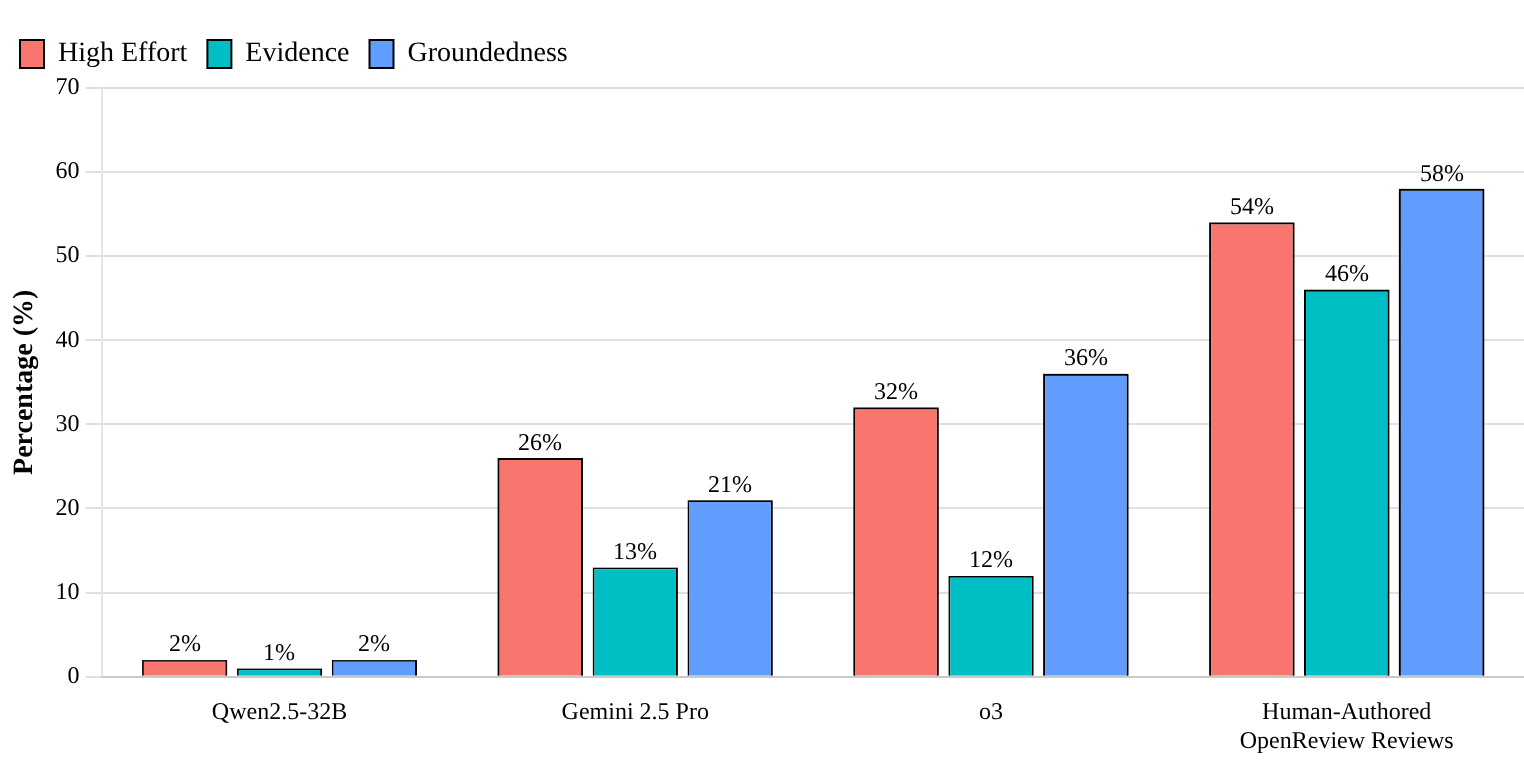}
  \caption{The figures show the distribution of votes on Effort, Evidence and Factual metrics for various sources of questions.}
  \label{fig:fig3votedistribution}
\end{figure}

\subsection{Examples of Questions Generated from  Openreviewer, DeepReviewer and IntelliAsk}

\label{examples}


\textbf{Paper 1:}\\
\textbf{Human-Authored}: Concerning the computation requirement, one gradient descent step seems to become at least slightly more computationally expensive as compared to DQN. In the paper it is mentioned at the very end of the discussion section that "with K=5 (it) only requires 1 to 2 times more time to run". Can this quantification be more accurate or why would it "sometimes" same compute time and "sometimes" double the compute time?

\textbf{IntelliAsk-32B}: The analysis in Section~5 uses Theorem~5.1 to bound the gap between the learned and optimal value functions, showing that iDQN's error depends on the sum of approximation errors over $K$ Bellman iterations. However, the bound also contains problem-dependent constants $C_{K,\gamma,R_\gamma}$ and $F(r; K, \rho, \gamma)$, whose scaling with $K$ is not explicitly characterized. How do these terms grow as $K$ increases, and what does this imply for the trade-off between reduced approximation error and potential amplification of concentrability coefficients or other problem-dependent factors? In particular, does increasing $K$ asymptotically improve the bound, or does it introduce a non-trivial optimization trade-off when choosing $K$ in practice?

\textbf{OpenReviewer}:
Why did you not compare to other DQN variants?

\textbf{DeepReviwer:}
Could the authors provide a clear explanation of the proposed algorithm, and how it compares to existing algorithms?





*The weights for AutoRev aren't open-sourced so we referred to the questions presented in the paper for evaluating the quality of questions.

\subsection{Likert Scoring Analysis}
\label{sec:likert_observations}

\begin{figure}[h]
    \centering
    \includegraphics[width=\linewidth]{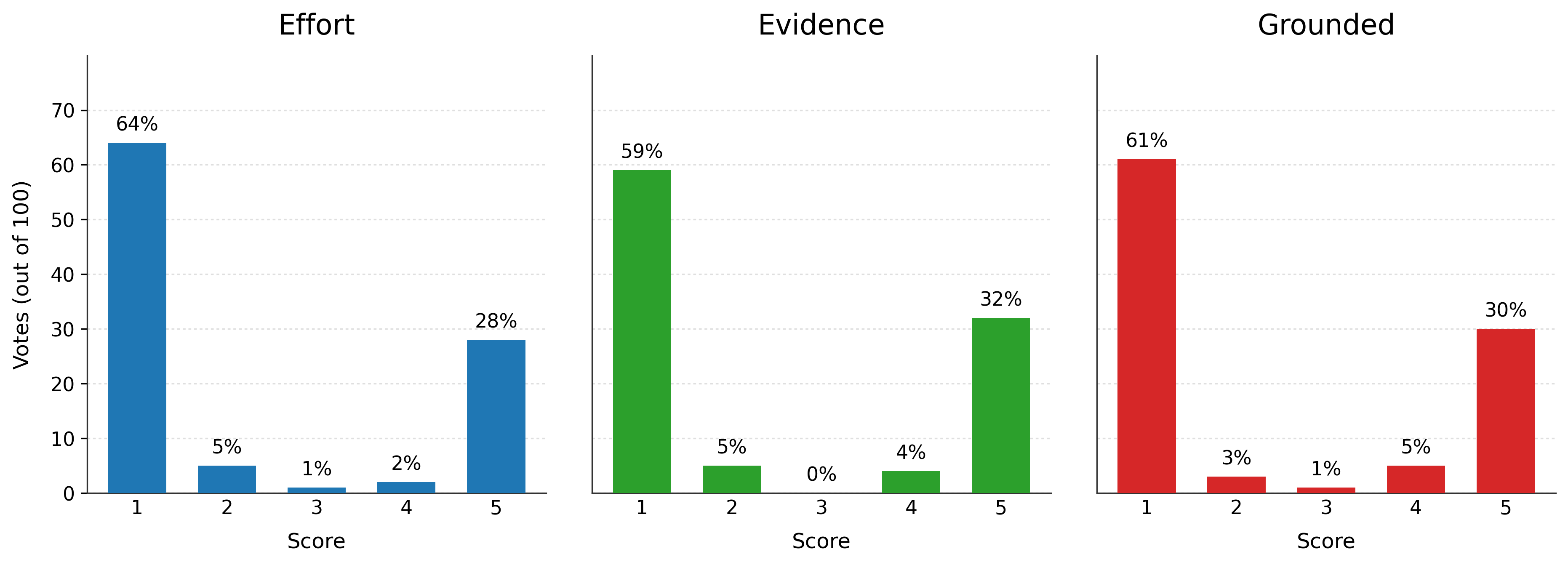}
    \caption{Distribution of votes across categories during pilot annotation. The data exhibits a clear clustering at the extremes (1 and 5).}
    \label{fig:pilotannotation}
\end{figure}

Initially, we explored a Likert scoring mechanism. During the pilot phase, annotators employed a 1--5 scale to evaluate Effort, Evidence, and Grounding. Upon completing 25\% of the annotations, however, we observed a strong bimodal pattern. As illustrated in Figure~\ref{fig:pilotannotation}, over 85\% of ratings clustered at the extremes (1 or 5), with sparse usage of intermediate values.

\subsection{Alignment of Reward Model with Human Judgments}
\label{sec:human_alignment}

\begin{figure}[h]
    \centering
    \includegraphics[width=\linewidth]{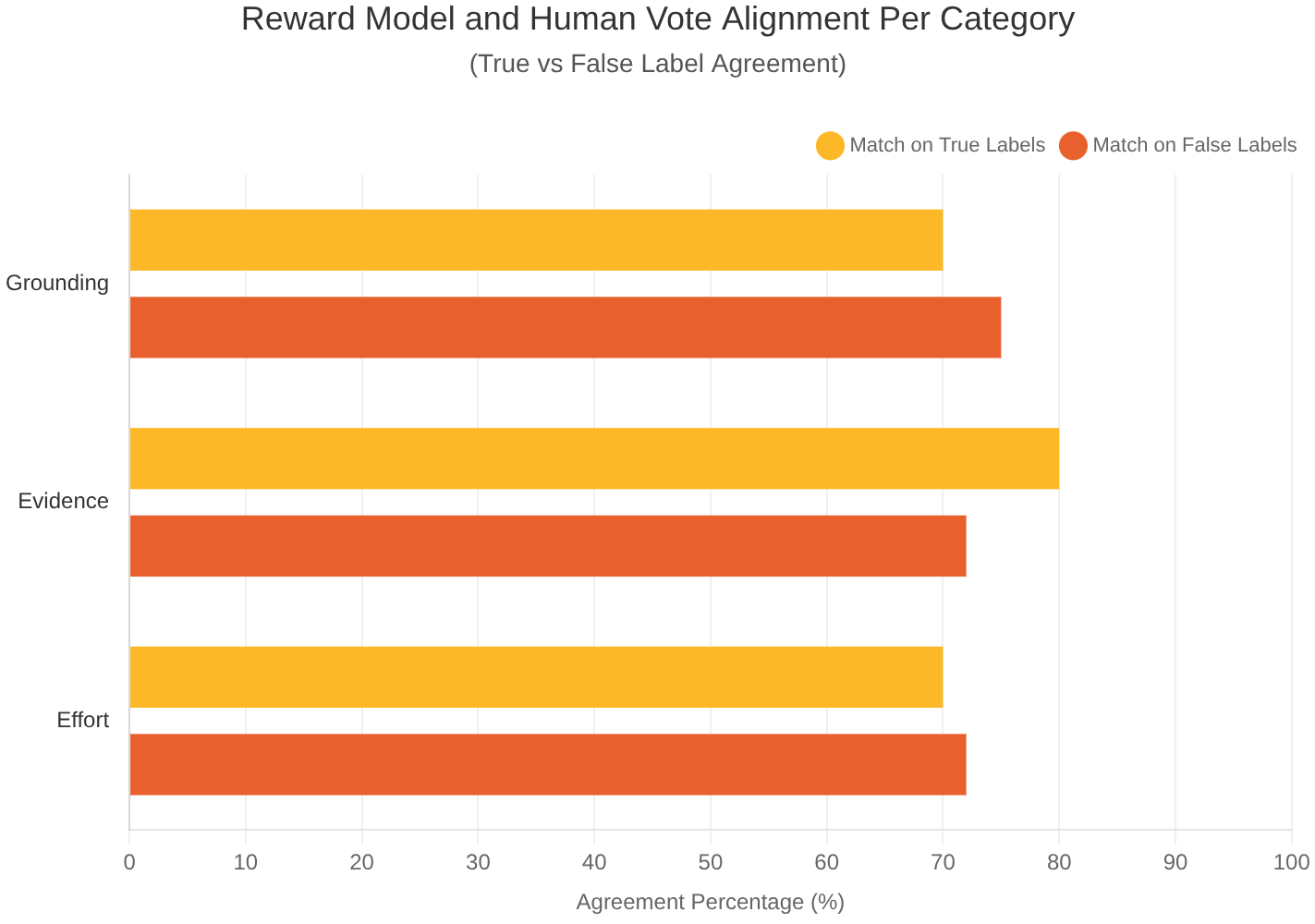}
    \caption{Agreement between the reward model and human annotations. The model achieves high consistency across Grounding, Evidence, and Effort for both positive and negative class labels.}
    \label{fig:humanalign}
\end{figure}

We evaluated the alignment between our reward model and human annotators across three key dimensions: Grounding, Evidence, and Effort. As shown in Figure~\ref{fig:humanalign}, the model demonstrates consistent agreement with human judgment, exceeding 70\% accuracy for both positive and negative labels across all categories.

\subsection{Rejection Sampling}
\label{sec:rejection_sampling}

Following the setup in \citet{webgpt}, we performed rejection sampling by generating 16 completions for each of the 300 prompts in the human preference annotation test set. We set the temperature to 0.9 and computed best-of-$n$ for $n \in \{1, 2, 4, 6, 8, 16\}$. Completions were generated using GPT-5 and Gemini-2.5-Pro.

\begin{figure}[H]
    \centering
    \includegraphics[width=\linewidth]{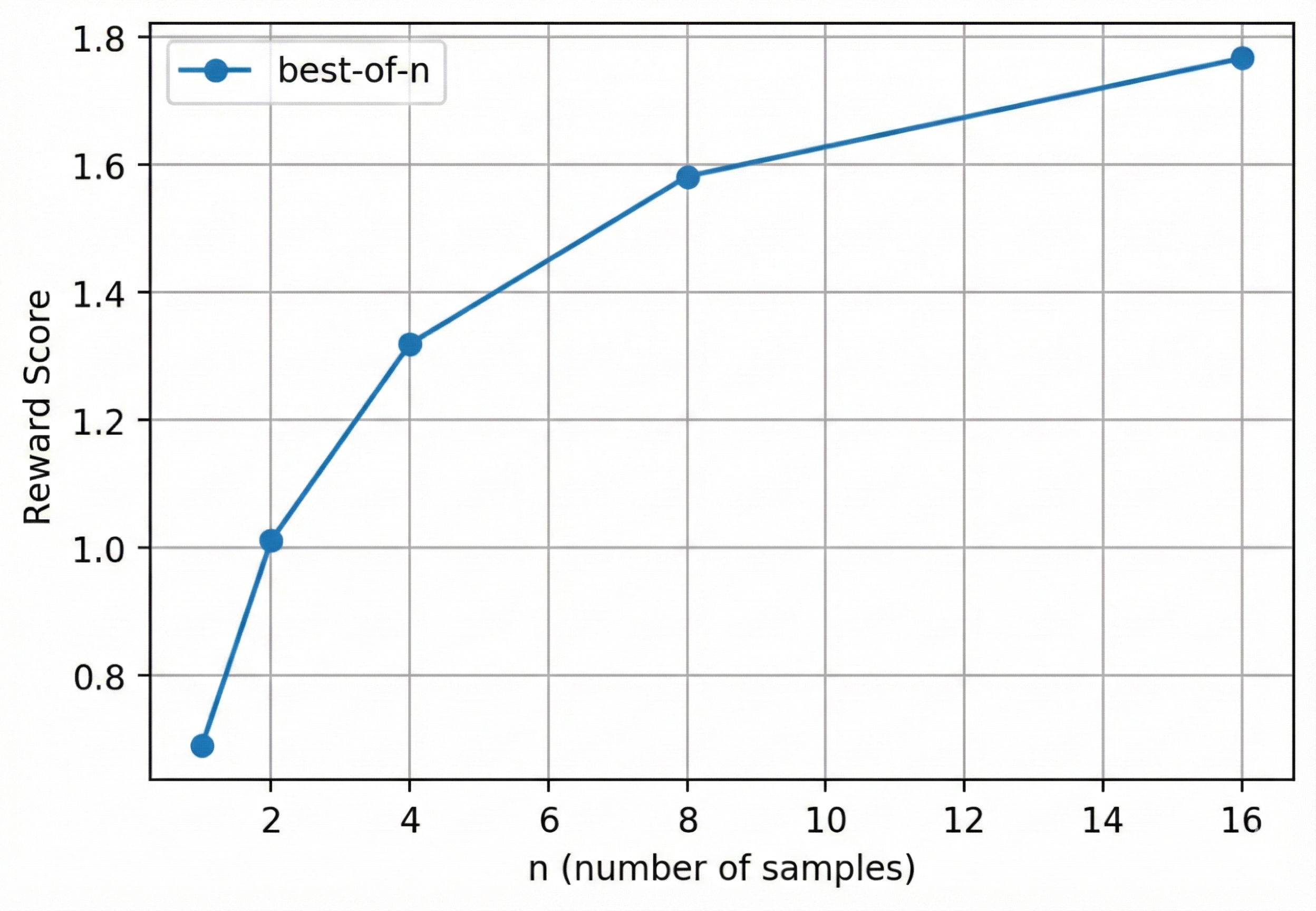}
    \caption{Reward Score using Best-of-$n$ for Gemini 2.5 Pro.}
    \label{fig:bestofn_gemini}
\end{figure}

\begin{figure}[H]
    \centering
    \includegraphics[width=\linewidth]{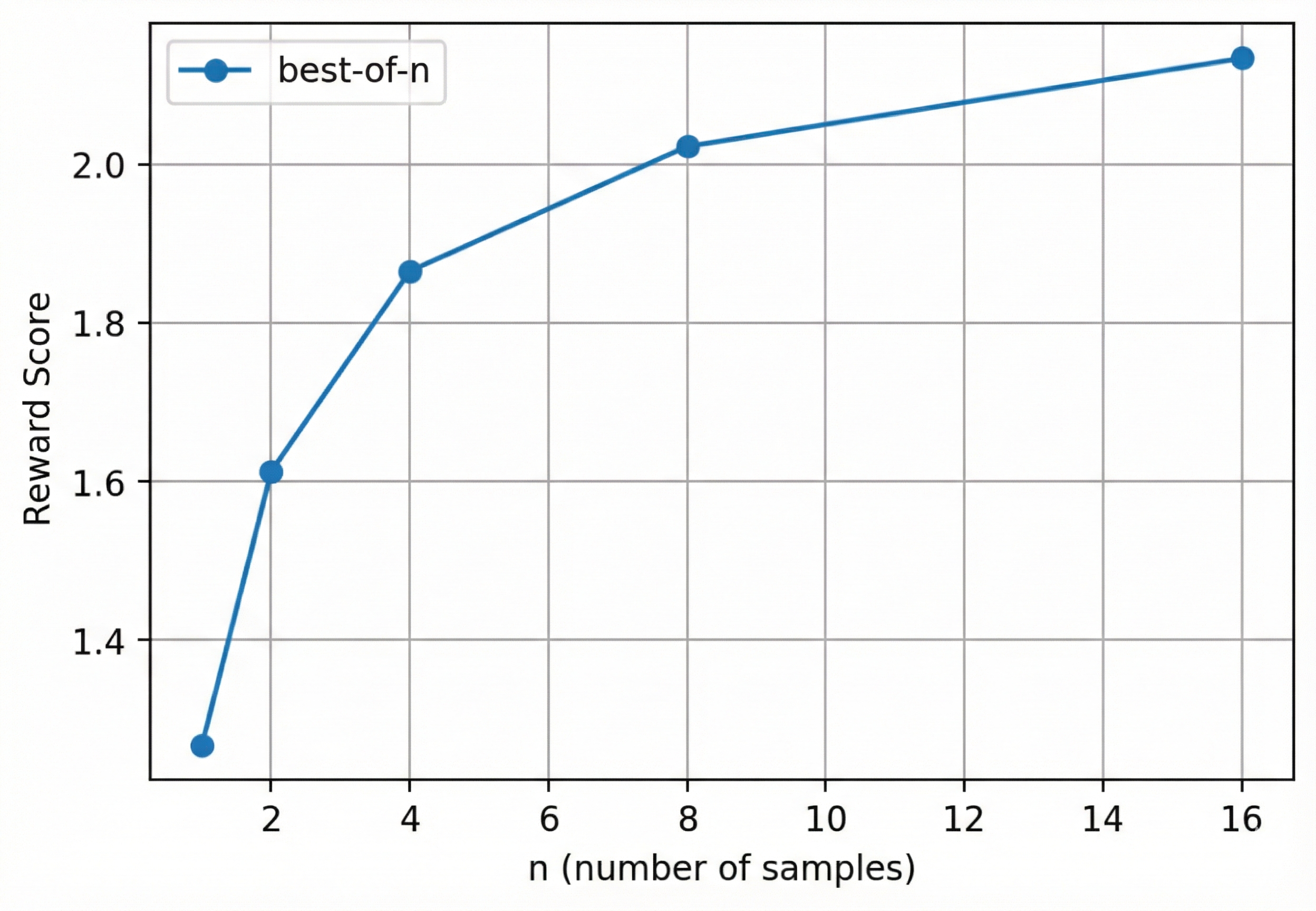}
    \caption{Reward Score using Best-of-$n$ for GPT-5.}
    \label{fig:bestofn_gpt5}
\end{figure}

Annotators manually inspected these samples to verify whether the reward scores matched the actual quality of the generated questions. We present selected examples from this analysis in Table~\ref{tab:rs}. Additionally, we summarize the best-of-$n$ results for both models in Table~\ref{tab:best_of_n_scaling} and illustrate the expected reward curves in Figure~\ref{fig:bestofn_gemini} and Figure~\ref{fig:bestofn_gpt5}.

\begin{table}[h] 
    \centering
    \small
    \renewcommand{\arraystretch}{1.3} 
    
    \begin{tabularx}{\columnwidth}{@{} X c @{}}
        \toprule
        \textbf{Question} & \textbf{Score} \\
        \midrule
        
        \rowcolor{gray!15} \multicolumn{2}{@{}l@{}}{\textbf{GPT-5}} \\ 
        In Algorithm 1, Eq. (2) appears to subtract identical terms at $x_{t-1}$; was the intended SPIDER-style recursion $u_t^s = u_{t-1}^s + (1/|A|) \sum_{j\in A}[\nabla f_{sj}(x_t; \xi_{sj}) - \nabla f_{sj}(x_{t-1}; \xi_{sj})]$, and if so, can you show why this estimator yields an unbiased $\lambda_t$-weighted common descent direction? & 3.0 \\
        \cmidrule(lr){1-2} 
        Why is permutation invariance inappropriate for Event Cloud processing, and how do PEPNet's tailored hierarchical structure with temporal attention aggregation achieve state-of-the-art relocalization accuracy? & 0.0 \\
        
        \addlinespace[0.5em] 
        
        \rowcolor{gray!15} \multicolumn{2}{@{}l@{}}{\textbf{Gemini 2.5 Pro}} \\
        How does the paper's decomposition of the Bayes-Adaptive MDP's Q-value into an `Incremental Value of Information' and a `Value of Opportunity' explain why different classes of reward shaping functions are effective? & 2.0 \\
        \cmidrule(lr){1-2} 
        How does the proposed framework enhance the robustness of reinforcement learning agents against adversarial state perturbation-inference techniques tailored for different types of environments? & 0.0 \\
        
        \bottomrule
    \end{tabularx}
    \caption{Qualitative comparison of generated questions. Gray headers indicate the model source. Scores reflect the reward model's evaluation of the generated text.}
    \label{tab:rs} 
\end{table}
\begin{table}[h]
    \centering
    \small 
    
    \renewcommand{\arraystretch}{1.3} 
    \setlength{\tabcolsep}{10pt} 
    
    \begin{tabular}{
      c 
      S[table-format=1.4] 
      S[table-format=+1.4] 
      S[table-format=1.4] 
      S[table-format=+1.4]
    }
    \toprule
    & \multicolumn{2}{c}{\textbf{Gemini 2.5 Pro}} & \multicolumn{2}{c}{\textbf{GPT-5}} \\
    \cmidrule(lr){2-3} \cmidrule(lr){4-5}
    {$n$} & {\text{Reward}} & {\text{Gain}} & {\text{Reward}} & {\text{Gain}} \\ 
    \midrule
    1  & 0.6896 & {---}    & 1.2667 & {---} \\
    2  & 1.0114 & +0.3218 & 1.6125 & +0.3458 \\
    4  & 1.3192 & +0.6296 & 1.8649 & +0.5982 \\
    8  & 1.5816 & +0.8920 & 2.0222 & +0.7555 \\
    16 & 1.7667 & +1.0771 & 2.1333 & +0.8667 \\
    \bottomrule
    \end{tabular}
    \caption{Best-of-$n$ Performance: Gemini 2.5 Pro vs. GPT-5}
    \label{tab:best_of_n_scaling}
\end{table}

    
    
    

\subsection{Question Preference: IntelliAsk-32B vs GPT-4.1, Gemini-2.5 Flash, Qwen3-32B}
\label{sec:pairwise_pref}

\begin{figure}[H]
    \centering
    \includegraphics[width=\linewidth]{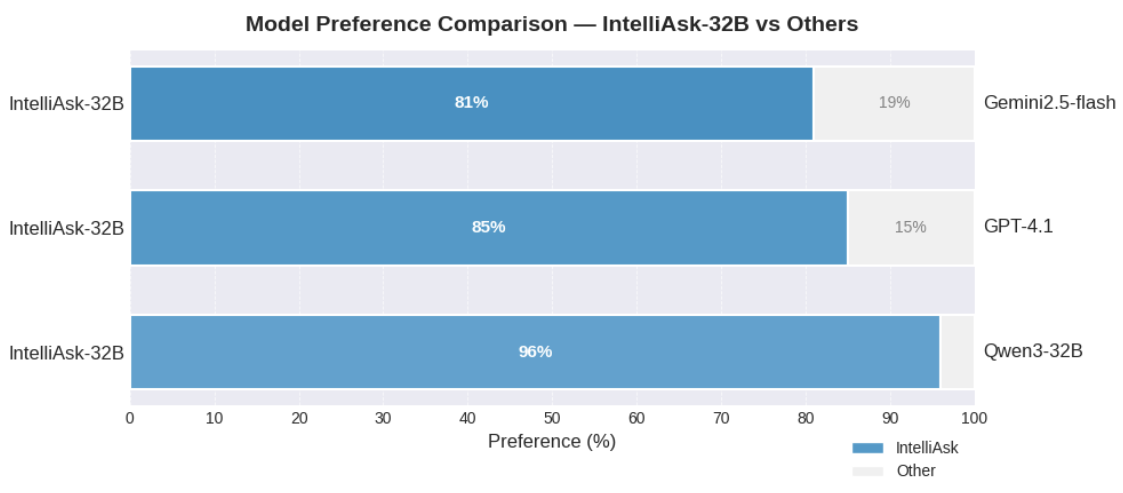}
    \caption{Pairwise preference results. IntelliAsk-32B is consistently favored over Gemini-2.5-Flash, GPT-4.1, and Qwen3-32B, receiving 81--96\% of total votes.}
    \label{fig:pref}
\end{figure}

To assess model quality, we conducted pairwise human preference evaluations, comparing IntelliAsk-32B against three strong baselines: Gemini 2.5-Flash, GPT-4.1, and Qwen3-32B. As shown in Figure~\ref{fig:pref}, IntelliAsk-32B achieved significantly higher preference rates across all comparisons, winning between 81\% and 96\% of evaluated pairs. These results underscore the model's substantial advantage in alignment with human judgment.

\subsection{Inter-Annotator Agreement}
\label{sec:inter_annotator}

\begin{figure}[h]
    \centering
    \includegraphics[width=\linewidth]{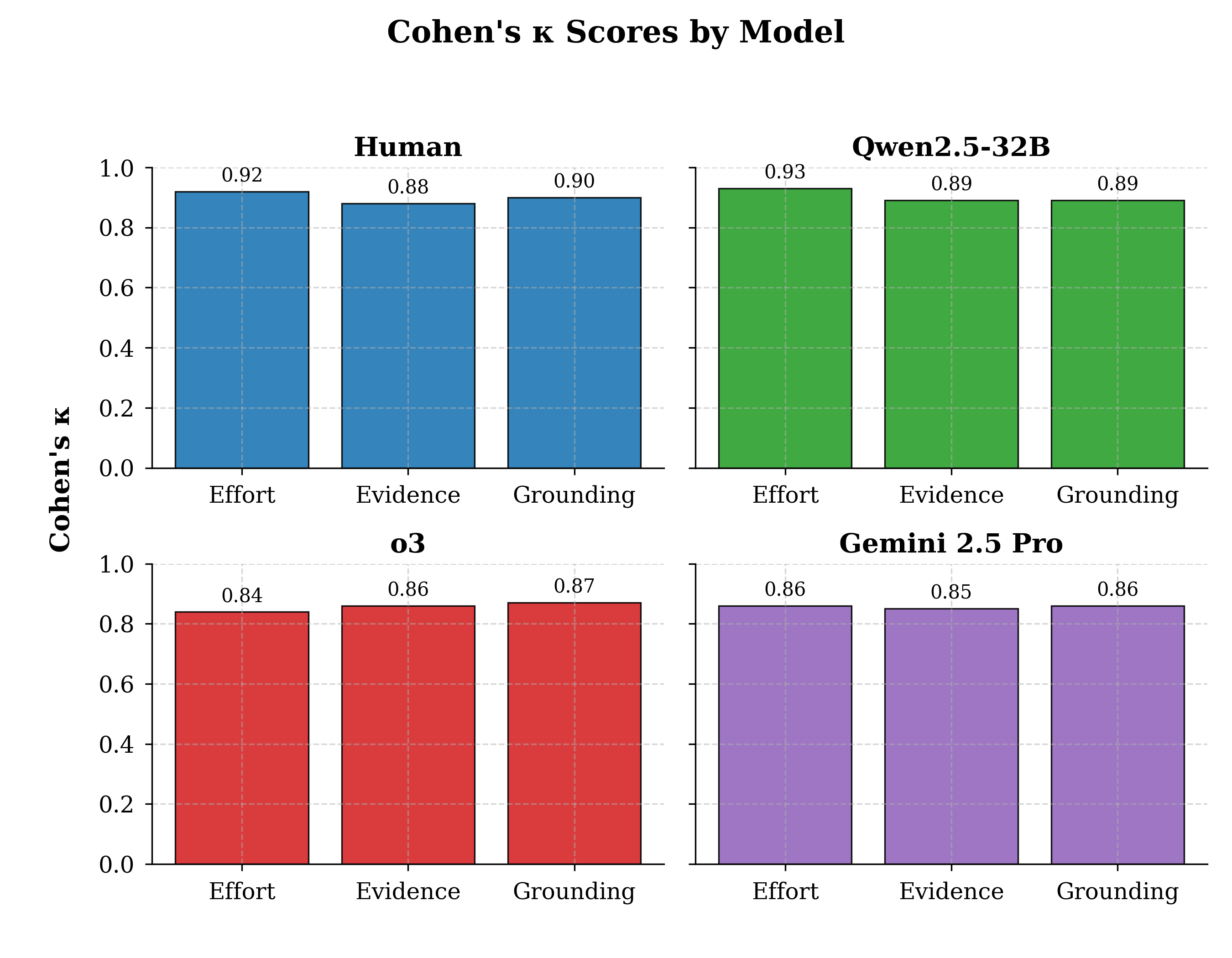}
    \caption{Cohen's $\kappa$ agreement scores. The results indicate consistent reliability across the Effort, Evidence, and Grounding evaluation categories.}
    \label{fig:cohen}
\end{figure}

We evaluated the reliability of our annotation process using inter-annotator agreement metrics on the human preference annotation data. Annotators demonstrated stable consistency across the three core attributes: Effort, Evidence, and Grounding. Figure~\ref{fig:cohen} presents the Cohen's $\kappa$ scores for each source, confirming high agreement levels.

\subsection{Score Distribution in WritingBench}
\label{sec:wb_scores}

Table~\ref{tab:comparison_scores} provides a detailed breakdown of the score distribution for IntelliAsk-32B and Qwen3-32B on WritingBench. The results indicate that \textbf{IntelliAsk-32B} demonstrates dominant performance, surpassing Qwen3-32B in the vast majority of evaluated domains and document categories.

\begin{table*}[t]
    \centering
    \small
    \renewcommand{\arraystretch}{1.1} 
    \setlength{\tabcolsep}{5pt} 
    
    \begin{tabular}{@{} lcc @{\hskip 0.5in} lcc @{}} 
        \toprule
        \textbf{Category} & \textbf{IntelliAsk} & \textbf{Qwen3} & \textbf{Category} & \textbf{IntelliAsk} & \textbf{Qwen3} \\
        \cmidrule(r){1-3} \cmidrule(l){4-6}
        
        Academic \& Engineering & \textbf{8.33} & 8.09 & Contract & \textbf{8.16} & 7.94 \\
        Finance \& Business & \textbf{8.22} & 8.04 & Test Report & \textbf{8.35} & 8.01 \\
        Politics \& Law & \textbf{8.29} & 8.02 & User Research & \textbf{7.93} & 7.72 \\
        Literature \& Arts & \textbf{8.41} & 8.16 & Meeting Minutes & \textbf{8.40} & 8.31 \\
        Education & \textbf{8.27} & 8.09 & Briefing & \textbf{8.37} & 8.05 \\
        Advertising \& Marketing & \textbf{8.37} & 8.18 & Financial Reports & \textbf{7.97} & 7.79 \\
        Abstract & \textbf{8.00} & 7.95 & Tender Document & \textbf{8.18} & 7.99 \\
        Introduction & \textbf{8.00} & 7.85 & Bid Proposal & \textbf{8.26} & 7.76 \\
        Contributions & \textbf{8.67} & 8.34 & Requirements Spec. & \textbf{8.45} & 8.35 \\
        Limitations & \textbf{8.36} & 8.17 & Product Proposal & \textbf{8.31} & 8.18 \\
        Conclusion & \textbf{8.60} & 8.26 & Investment Analysis & 8.18 & \textbf{8.21} \\
        Literature Review & 8.30 & \textbf{8.31} & Risk Management & 8.17 & \textbf{8.18} \\
        Experiments & \textbf{8.53} & 8.11 & Market Analysis & 7.96 & \textbf{8.11} \\
        Defense Presentation & \textbf{7.93} & 7.75 & Human Resource Mgmt & \textbf{8.40} & 8.24 \\
        Defense Script & \textbf{7.96} & 7.74 & Market Research & \textbf{8.40} & 8.31 \\
        Technical Doc. & \textbf{8.45} & 8.31 & Recruitment & \textbf{8.30} & 8.21 \\
        Research Proposal & \textbf{8.33} & 7.82 & Pitch Deck & \textbf{8.43} & 8.18 \\
        Internship Report & \textbf{8.80} & 8.60 & Event Planning & \textbf{8.32} & 8.13 \\
        Engineering Report & \textbf{8.70} & 8.40 & Business Corresp. & \textbf{8.00} & 7.62 \\
        Patent & 8.30 & \textbf{8.31} & Party Membership App & \textbf{9.00} & 8.75 \\
        
        \midrule
        \rowcolor{gray!20}
        \multicolumn{6}{c}{\textbf{Overall Mean Score: \quad IntelliAsk-32B = 8.31 \quad vs \quad Qwen3-32B = 8.07}} \\
        \bottomrule
    \end{tabular}
    \caption{Detailed Performance Comparison by Domain (Score out of 10) on WritingBench. The list is split into two columns for compactness. \textbf{IntelliAsk-32B} (ours) consistently outperforms Qwen3-32B.}
    \label{tab:comparison_scores}

\end{table*}

\subsection{TRAIT Benchmark Analysis}
\label{sec:trait_analysis}

Table~\ref{tab:trait_benchmark} presents the results of the TRAIT benchmark. The metrics are divided into the Big Five personality traits and the Dark Triad. For \textit{Neuroticism} and the \textit{Dark Triad}, lower scores indicate safer, more aligned behavior. IntelliAsk-32B demonstrates significantly lower (safer) scores across these negative dimensions compared to Qwen3-32B.

\begin{table}[H]
    \centering
    \small
    \renewcommand{\arraystretch}{1.2}
    
    \begin{tabular}{@{} l c c @{}}
        \toprule
        \textbf{Trait} & \textbf{IntelliAsk-32B} & \textbf{Qwen3-32B} \\
        \midrule
        
        \rowcolor{gray!15} \multicolumn{3}{@{}l}{\textit{\textbf{Big Five (Positive)}}} \\
        Openness            & \textbf{0.679} & 0.611 \\
        Conscientiousness   & 0.714 & \textbf{0.754} \\
        Extraversion        & 0.364 & \textbf{0.485} \\
        Agreeableness       & 0.667 & \textbf{0.781} \\
        
        \addlinespace[0.5em]
        \rowcolor{gray!15} \multicolumn{3}{@{}l}{\textit{\textbf{Negative Traits (Lower is Better)}}} \\
        Neuroticism         & \textbf{0.160} & 0.209 \\
        Machiavellianism    & \textbf{0.115} & 0.258 \\
        Narcissism          & \textbf{0.105} & 0.115 \\
        Psychopathy         & \textbf{0.000} & 0.016 \\
        
        \bottomrule
    \end{tabular}
    \caption{Personality trait comparison. Bold values indicate the preferred result (Higher is better for positive traits; Lower is better for Neuroticism/Dark Triad).}
    \label{tab:trait_benchmark}
\end{table}

\subsection{Question Placement in Reviews}
\label{sec:question_placement}

\begin{figure*}[h] 
    \centering
    \includegraphics[width=\textwidth]{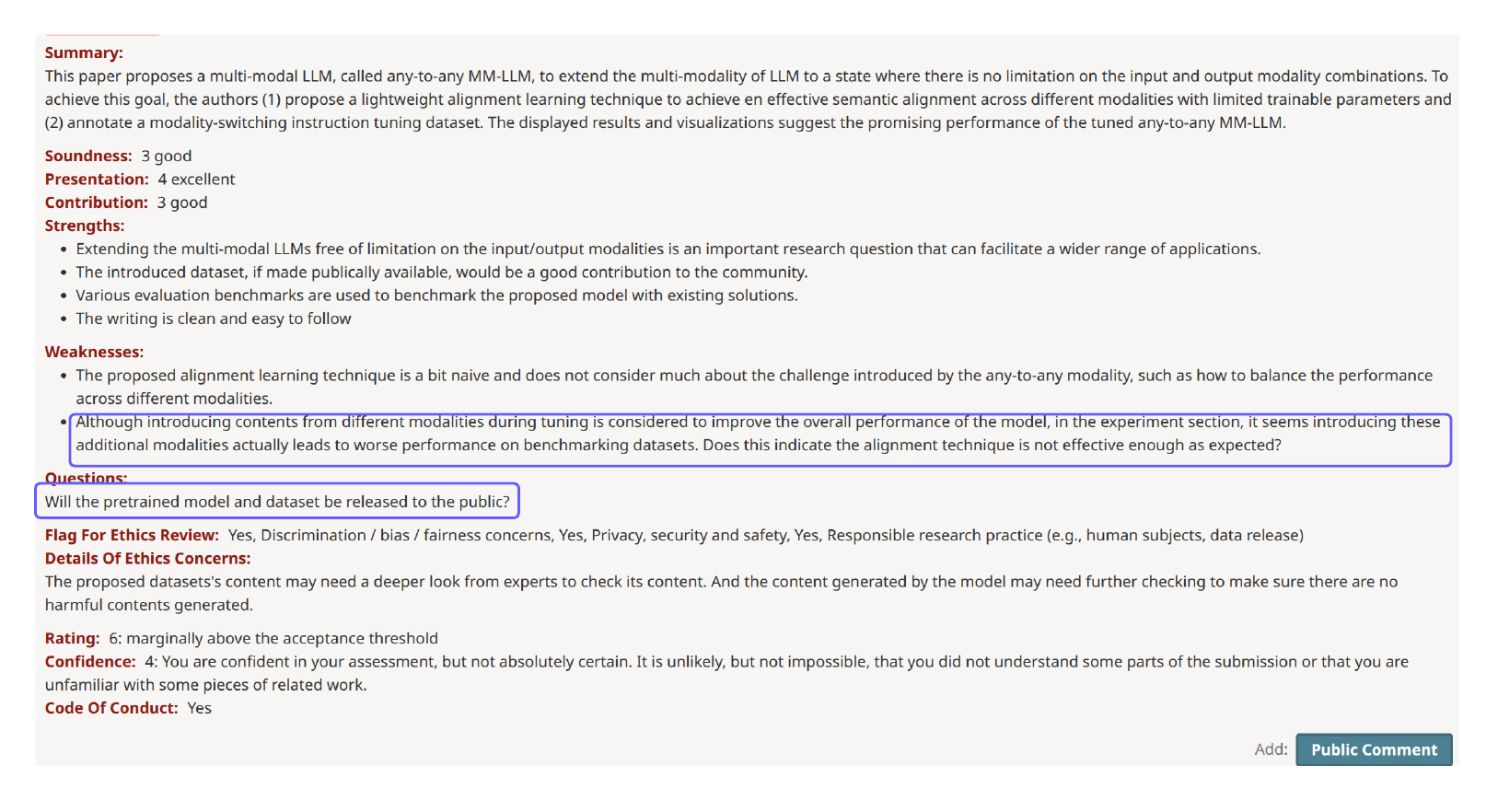}
    \caption{Variability in the occurrence of questions within reviews.}
    \label{fig:question_dist}
\end{figure*}

Figure~\ref{fig:question_dist} illustrates the positional distribution of questions within review texts. This highlights the variability of where questions might occur.

\subsection{Examples of Effortful, Substantive, and Evidence-Based Questions}
\label{sec:samples_rubric}

Table~\ref{tab:question_analysis} provides concrete examples distinguishing high and low quality across our three core dimensions: Effort, Evidence, and Grounding.

\begin{table*}[h]
    \centering
    \small 
    \caption{Analysis of Peer Review Questions. We contrast high-scoring versus low-scoring questions across three dimensions. \textbf{Q:} denotes the Question, and \textit{Reasoning} explains the score.}
    \label{tab:question_analysis}
    \renewcommand{\arraystretch}{1.3}
    
    \begin{tabularx}{\textwidth}{@{} p{1.8cm} X X @{}}
        \toprule
        \textbf{Dimension} & \textbf{High Quality Example} & \textbf{Low Quality Example} \\
        \midrule
        
        \textbf{Effort} & 
        \textbf{Q:} Why is the training time of NoLA with shared random basis similar to that of LoRA when the training time of NoLA with a unique random basis is higher? Aren't the number of coefficients being trained the same in both cases? \newline
        \textit{\textbf{Reasoning:} This requires reasoning about subtle implementation details and connecting training dynamics to design choices not explicitly stated in the paper.} & 
        \textbf{Q:} How does the proposed $\Delta$-SGD method adapt to the heterogeneity in local data across different clients and datasets compared to other optimization methods as shown in the experimental results? \newline
        \textit{\textbf{Reasoning:} The abstract and results already explicitly explain this. The answer requires only surface-level restatement without synthesis.} \\
        \cmidrule(lr){1-3}
        
        \textbf{Evidence} & 
        \textbf{Q:} `This way, we transform... bypassing the time-consuming gradient computation...' --- For MINE, we do need to update NNs' parameters. But InfoNet also needs gradient ascent. How to understand `bypassing the time-consuming gradient computation'? \newline
        \textit{\textbf{Reasoning:} Cites a specific claim to challenge a potential inconsistency. The critique is precise and grounded in the author's own text.} & 
        \textbf{Q:} What specific improvements or changes in the recommendation system's architecture or methodology did the authors implement to achieve improved performance compared to traditional systems? \newline
        \textit{\textbf{Reasoning:} Asks broadly about improvements without pointing to any specific claim, experiment, or section. Lacks evidence-based grounding.} \\
        \cmidrule(lr){1-3}
        
        \textbf{Grounding} & 
        \textbf{Q:} In section 4.2 you mentioned that you used LoRA to inject low-rank matrices into attention weights Q, K and V only... what is the rationale of only applying LoRA to Q, K and V? \newline
        \textit{\textbf{Reasoning:} Explicitly refers to Section 4.2 and concrete implementation choices, probing a decision directly anchored in the text.} & 
        \textbf{Q:} How does the proposed DRL framework address the trade-off between minimizing taxi delays and ensuring throughput... and how does this compare to Ali et al. (2022)? \newline
        \textit{\textbf{Reasoning:} The comparison is generic and does not engage with specific method details. The reference is already in the paper; the question adds no new depth.} \\
        
        \bottomrule
    \end{tabularx}
\end{table*}

\subsection{Training Configuration}
\label{sec:hyperparameters}

We utilized the \textit{grpo} estimator for adversarial training. The specific hyperparameters used for training IntelliAsk are detailed in Table~\ref{tab:hyperparams}.

\begin{table}[H]
    \centering
    \caption{Training parameters for IntelliAsk-7B}
    \label{tab:hyperparams}
    \small
    \setlength{\tabcolsep}{8pt}
    \renewcommand{\arraystretch}{1.1}
    
    \begin{tabular}{@{} l l @{}}
        \toprule
        \textbf{Parameter} & \textbf{Value} \\
        \midrule
        \multicolumn{2}{@{}l}{\textit{Experiment Metadata}} \\
        Model & Qwen2.5-7B-Instruct-1M \\
        Estimator & DAPO \\
        \addlinespace[0.5em]

        \multicolumn{2}{@{}l}{\textit{Core Training}} \\
        Clip Ratio & 0.20 (low) -- 0.28 (high) \\
        Max Prompt Length & 14,000 \\
        Max Response Length & 20,480 \\
        Overlong Buffer & Enabled (Length: 15,024) \\
        Loss Aggregation Mode & token-mean \\
        Filter Groups Metric & acc (Enabled) \\
        \addlinespace[0.5em]
        
        \multicolumn{2}{@{}l}{\textit{Batch Sizes}} \\
        Max Num Gen Batches & 2 \\
        Train Prompt Batch Size & 64 \\
        Gen Prompt Batch Size & 192 \\
        Responses per Prompt & 8 \\
        Train Prompt Mini Batch & 2 \\
        Use Dynamic Batch Size & True \\
        \addlinespace[0.5em]

        \multicolumn{2}{@{}l}{\textit{Optimizer \& Actor}} \\
        Learning Rate & $1\text{e-}6$ \\
        Warmup Steps & 10 \\
        Weight Decay & 0.1 \\
        Entropy Coeff & 0.0 \\
        Grad Clip & 1.0 \\
        Temperature & 1.0 \\
        Top-p & 1.0 (Train), 0.7 (Val) \\
        \bottomrule
    \end{tabular}
\end{table}




\subsection{Annotation Interface}
\label{sec:annotation_ui}
\begin{figure*}[h]
    \centering
    \includegraphics[width=\textwidth]{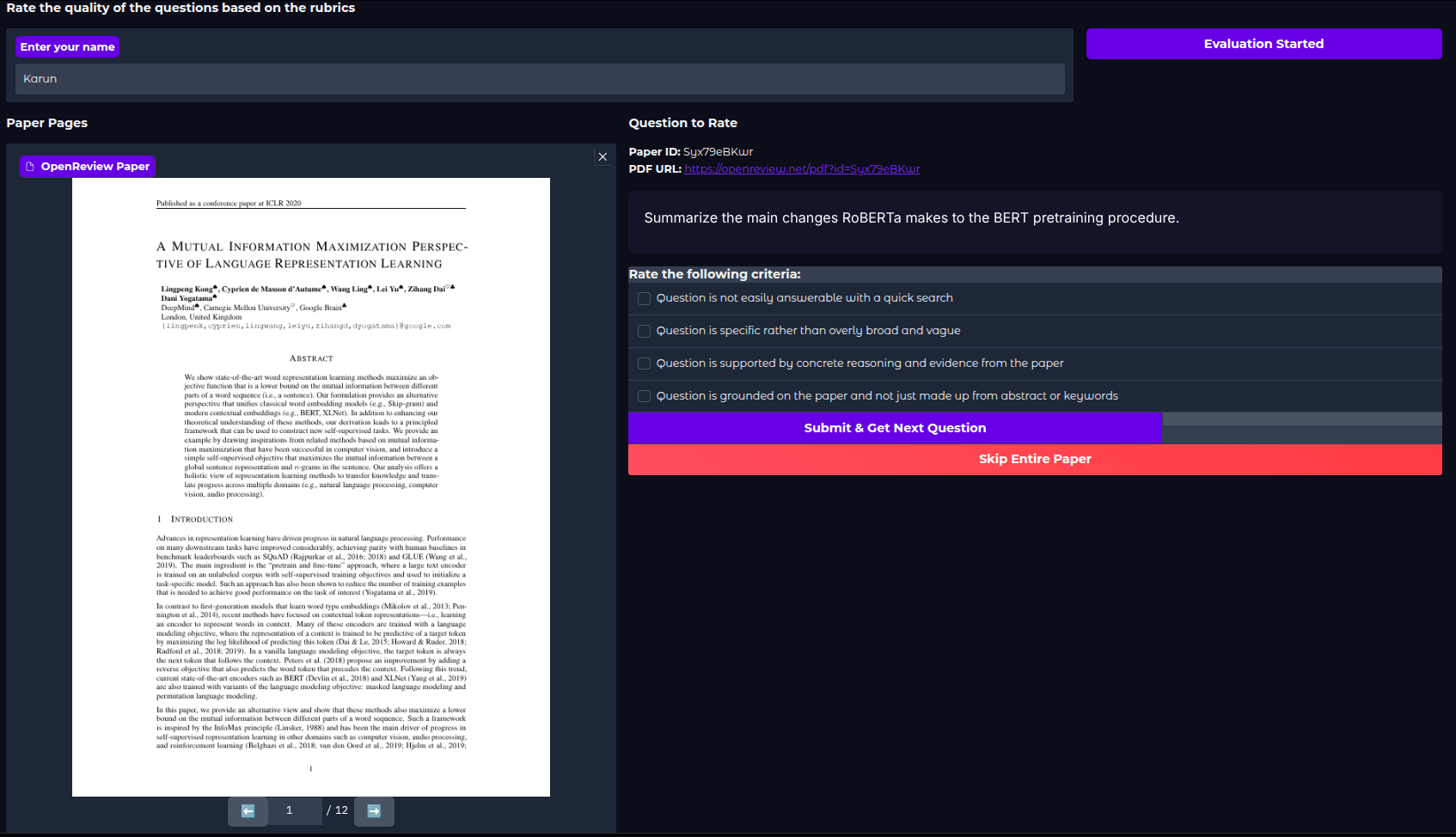}
    \caption{User Interface of the Human Annotation Tool. The screenshot demonstrates the layout used by annotators to grade model outputs (dummy data shown for illustration).}
    \label{fig:ui_screenshot}
\end{figure*}

No external annotators, crowdworkers, or paid participants were used. As paper authors conducting their own research, no compensation was provided for annotation work. 
Figure~\ref{fig:ui_screenshot} displays the user interface employed for human annotation. The tool was designed to streamline the evaluation of Effort, Evidence, and Grounding.

\onecolumn 
\raggedbottom
\subsection{System Prompt}
\subsubsection{Quality Gate 3}
\label{qg3}
\label{fig:prompt_qg3}

\begin{lstlisting}[style=promptstyle, caption={System Prompt for Quality Gate 3}]
You are an expert evaluator assessing Questions asked by the reviewers at top conferences from the CVPR, NeurIPS, ICML, ICLR, EMNLP, after reading a scientific paper  for their suitability in a specialized dataset aimed at training Large Language Models for advanced reasoning.

**Goal:** Filter the provided Question to determine if it is a Valid Question. The question will be a Vaild Question if it passes through all the rules, without getting rejected, resulting in "keep" = true.

**Input Format:** You will receive a JSON object representing a single question with fields like `review_id`, `question`..

**Output Format:** Respond with a JSON object containing two fields:
1.  `keep`: A boolean value (`true` or `false`).
2.  `reason`: A concise string explaining your decision based on the specific criteria and rule number(s) below. (e.g., "REJECT: Rule 2- Question states to correct the caption.", "KEEP: A Valid Question passed through all the rules.").

**Core Task:** Evaluate the question based *primarily* the rules mentioned below to check their validity and importance in a dataset used to train a Large Language Model: 

**Filtering Criteria & Rules (Apply strictly in this order):**
**Rule 1**: REJECT the questions asking for changes/additions/formatting that require substantial effort  
**Rule 2**: REJECT the questions asking for Edits, Summaries, correcting typos  
Examples of Questions to REJECT under this rule: 
Question: In Table 2, it probably needs to be noticed that for COCO instance segmentation, Mask R-CNN is used
 Question: Correct the typo made on page 4, line 3 and add a caption for figure 3. 
**Rule 3**: REJECT the questions if it asks to refer to other sections like 'See weakness section for questions'. 
**Rule 4**: REJECT the questions if it contains unprofessional or inappropriate remarks in the review and giving personal opinions on the paper quality
 	Examples of Questions to REJECT under this rule: 
Question: I spend several hours and still can not get an intuitive understanding about why such a claim hold. For instance, why A and B are 'irrelevant' according to footnote 6?

Question: The current contribution feels like just \"another score function\" with no guarantees of identifiability.

Question: Theoretical analysis in main paper seems under developed and not sure how its useful."


**Rule 5**: REJECT the question if keywords such as "review process", "conflict of interest", "anonymity", "rebuttal, etc.. appear.


**Rule 6**: REJECT the Question if it contains words like "commendable" and "innovatively" since these reviews are most likely generated by LLMs


**Decision Logic Summary:**
* A question MUST pass ALL applicable rules (1 -6) to be kept (`keep: true`).
* Failure at any rule stage leads to rejection (`keep: false`).
\end{lstlisting}

\subsubsection{Quality Gate 4}
\label{qg4}

\begin{lstlisting}[style=promptstyle, caption={System Prompt for Quality Gate 4}]
You are an expert evaluator assessing Questions asked by the reviewers at top conferences from the CVPR, NeurIPS, ICML, ICLR, EMNLP, after reading a scientific paper for their suitability in a specialized dataset aimed at training Large Language Models for advanced reasoning.

**Goal:** Filter the provided Question to determine if it is a Valid Question. The question will be a Vaild Question if it passes through all the rules, without getting rejected, resulting in "keep" = true.

**Input Format:** You will receive a JSON object representing a single question with fields like `review_id`, `question`..

**Output Format:** Respond with a JSON object containing two fields:
1.  `keep`: A boolean value (`true` or `false`).
2.  `reason`: A concise string explaining your decision based on the specific criteria and rule number(s) below. (e.g., "REJECT: Rule 2- Question states to correct the caption.", "KEEP: A Valid Question passed through all the rules.").

**Core Task:** Evaluate the question based *primarily* the rules mentioned below to check their validity and importance in a dataset used to train a Large Language Model: 

**Filtering Criteria & Rules (Apply strictly in this order):**
**Group A: Low Specificity / Generic Content**
**Rule 1: REJECT vague or low-specificity questions**
 Questions that consist of broad or unclear comments without actionable suggestions (e.g.,  "Can you elaborate on the methodology?") should be REJECTED.


**Rule 2: REJECT generic questions about limitations or future work**
 REJECT questions that ask casually about limitations or future directions without referencing a specific issue, weakness, or observation in the paper.
          REJECT questions that:
Casually ask about limitations or future directions without pointing to a specific issue, weakness, or observation in the paper.
Use broad or vague phrasing like "Can you discuss the limitations...", "How could future work address this...", or "What are the next steps?" without context or justification.
 
 Examples of Questions to REJECT under this rule: 
Question: Can you discuss the limitations of your benchmarking tool, and how future research could address these limitations to further advance the field of PINNs
Only keep such questions if they are tied to concrete findings, results, or gaps explicitly discussed in the paper.


**Rule 3: REJECT superficial or generic feedback**
 REJECT out comments that offer only brief praise or criticism without actionable insight. Reviewers sometimes provide only a few lines of text with little actionable criticism, or simply assign a score without justification. This is irrelevant and low quality
 	Examples of Questions to REJECT under this rule: 
: "Great work!" with no follow-up question.


: "Writing too bad" or "not state of the art" or "too niche" etc.. without justification.


 
**Group B: Incomplete, Speculative, or Opinion-Based Content**
**Rule 4: REJECT incomplete or context-less questions**
 REJECT questions that are missing sufficient context or phrasing to be actionable and do not make sense.


Example: "Not really large-scale."


Example: "Ablation studies are missing."
Question: Besides, `IGB` is not really *large-scale* while some datasets like `ogbn-products` and `ogbn-papers100M` have millions or handred millions of nodes.


**Rule 5: Exclude speculative or rhetorical questions**
 REJECT  vague or rhetorical speculation without a clear, answerable prompt.


Example: "I assume they come from different sources..."


Example: "Would this method fail if we used another model?"
Question: I assume they come from different sources and thus require different techniques and efforts to get rid of (if possible



**Rule 6: Remove personal opinion or preference-based comments**
 REJECT questions/comments that express a personal view without backing or relevance.


Example: "...which is not that necessary, in my opinion."


**Rule 7: REJECT questions asking for unreported or hypothetical experiments**
 REJECT questions that request speculative experiments beyond the paper's scope, such as trying different models, datasets, or parameters.
Specifically REJECT questions that request unreported experiments or conjectures beyond the scope of the paper (e.g., "Could this work better with another model?", "What happens if we try Z instead?").

Examples of Questions to REJECT under this rule: 
Question: Compared to Hits@10, Hits@1 could be more critical in the real-world applications, especially for tail nodes with very few neighbors. I wonder if the authors can also provide the Hits@1 performance.
Question: Would the method fail if using a non-contrastive pre-trained model?
The paper mainly focuses on 4-bit and 5-bit quantization, leaving questions about the performance and relevance of other bit quantizations



**Rule 8: Exclude questions framed as unsupported suggestions**
 REJECT questions like "Did you consider X?" if they are isolated and not grounded in the paper's content, especially if surrounded by uninformative praise or vague critique.




Make sure to be strict so that no poor quality question passes through.

**Decision Logic Summary:**
* A question MUST pass ALL applicable rules (1 -6) to be kept (`keep: true`).
* Failure at any rule stage leads to rejection (`keep: false`).
\end{lstlisting}


\subsubsection{Question Generation}
\label{qg}

\noindent The prompt shown below was used uniformly across all models for question generation.

\begin{lstlisting}[style=promptstyle, caption={Prompt for Question Generation}]
{"role": "system", "content": "You are expert at asking unique questions based on the OCR text of a research paper. So given the text, generate one high quality question now."},
{"role": "user", "content": f"Here's the text of the complete research paper and now generate a question based on it. \n{ocr_output}"}
\end{lstlisting}


\subsubsection{Extraction of Questions}
\label{extrac}

\begin{lstlisting}[style=promptstyle, caption={System Prompt for Question Extraction}]
"""You are a highly experienced professor from Stanford University with extensive experience in reviewing and publishing research papers. You will be provided with a peer review containing a heading called "Questions" and another section called "Mixed Content". The "Questions" section contains multiple questions without any indication/ separator for a new question and the "Mixed Content" has a mix of questions that might not have a "?" to indicate a question. It can simply be a suggestion, an edit, a clarification required from the author etc.


Task: Your Primary task is to Extract Questions first from the "Questions" section and then from the "Mixed Content" section. Perform verbatim extraction. I.e. Word-for-Word
By Questions I mean all the questions --- explicitly or implicitly asked --- that the author needs to answer the reviewer based on the review text.


1) Extract all the questions from the "Questions" section in a way all the sentences are retained. Do not miss any sentence or words from the original content in the section and output multiple Questions you have found, you need to break the Questions properly. If someone concatenates the multiple questions you have formed, they must get the "Questions" section as it is.
2) While breaking the questions from the "Question" section, you might encounter nested questions. If both the parts are related keep them as a single question but if one part is an independent question, make them as separate questions.
3) Extract all the questions that are present in the "Mixed Content" section. The questions might not be direct, it might include the reviewer telling what made him arrive at this question and then pose the question. It can also be some clarification he/she needs from a content in the paper. So include the complete context and don't simply output just the question.
4) In some cases, the "Questions" section will direct you to refer the "Mixed Content" section by asking you to refer the weakness. That simply is your hint to find questions in the "Mixed Content" section.
5) The "Mixed Content" section might have general observations or weaknesses of the paper, so only pick up questions,reviewer's suggestion for edits, reviewer seeking clarification BUT don't include general observations. This is the rule for "Mixed Content" section.


Note: The "Questions" section will always have question present in it until unless it is blank or only asking you to refer to the weakness. The "Mixed Content" section might or might not have questions in it, so check very carefully. Learn from the zero-shot example below.
Note 2: Important: When the questions that you form from "Questions" section are concatenated, it should form the original and complete content of the "Questions" section. This rule of concatenation is important and ONLY for "Questions" Section ONLY.

Remember: Your task is just extraction of Questions and Not Rephrasing.


###Output
Questions: [
{
\"Paper_id\": <ID of Paper>,
\"review_id\": <ID of review>,
\"Q_Number\": <Index of generated question>,
\"Question\": <Extracted question>
},
]




Example 1:


Input:


Paper Id: : Asdho34
Review Id: ioedh45
"Questions" :
"I have questions about the learning process of the 1x1 conv layer in equation (5). How is it exactly trained? And is it sensitive to the training sample size?\
- Will instance normalization also work in text-to-image tasks? It will be interesting to see if it could generate higher fidelity images with semantic meaning more aligned with the provided text prompts"


"Mixed Content" :
"The proposed method is a systematic approach for image translation tasks incorporating different components. A potential drawback is its inference speed. It would be beneficial if the authors could compare inference speed with other image translation tasks.\
- The comparison with methods like SDEdit, Prompt2Prompt, and InstructPix2Pix is somehow unfair since they do not require an additional segmentation network.\
- The quantitative evaluation is only the proposed dataset, which contains fine-grained edit instructions. The effectiveness of DVP could be further proved by evaluating simple or even ambiguous instructions


Overall, the paper is well-organized and easy to follow. The figures and tables are informative.\
\
- The performance of the proposed method is promising. Figures 4, 6 clearly demonstrate the superiority of DVP.\
\
- The ablation study and system analysis are clear and informative, making it easy to see the effectiveness of different parts, such as instance normalization, and prompte."




Output:
{
Questions: [
{
\"Paper_id\": Asdho34,
\"review_id\": ioedh45,
\"Q_Number\": 1,
\"Question\":  ""I have questions about the learning process of the 1x1 conv layer in equation (5). How is it exactly trained? And is it sensitive to the training sample size?"
},


{
\"Paper_id\": Asdho34,
\"review_id\": ioedh45,
\"Q_Number\": 2,
\"Question\":  "Will instance normalization also work in text-to-image tasks? It will be interesting to see if it could generate higher fidelity images with semantic meaning more aligned with the provided text prompts"
},
{
\"Paper_id\": Asdho34,
\"review_id\": ioedh45,
\"Q_Number\": 3,
\"Question\":  "A potential drawback is its inference speed. It would be beneficial if the authors could compare inference speed with other image translation tasks"
},


{
\"Paper_id\": Asdho34,
\"review_id\": ioedh45,
\"Q_Number\": 4,
\"Question\":  "The quantitative evaluation is only the proposed dataset, which contains fine-grained edit instructions. The effectiveness of DVP could be further proved by evaluating simple or even ambiguous instructions
"
}
]


}




Example 2


Input:


Paper Id: : Asdho34
Review Id: ioedh45
"Questions" :
"Please comment on the weaknesses outlined above.\
- Figures 10 and 11, right: Why is adaptation slower for OC-GFN than GFN in the first few thousand iterations? This is surprising since one would hope pretraining helps bootstrap downstream performance as in vision / language / RL. If it's an exploration phase, did you validate it and is there a way to side-step it?"


"Mixed Content" :
"There should be a discussions of assumptions behind the OC-GFNs pretraining. Namely, that transfer is only possible when the reward function changes but not if the action-space or the state-space change. Moreover, the goal-conditioning requires a well specified set of outcomes Y --- presumably not all states s are terminal states --- which makes the proposed method not truly unsupervised. These limitations (together with the applicability mentioned at the end of A.2) could be stated explicitly in the main text, and left to future work.\
- While there are enough benchmarks, I believe none include continuous action/state spaces. Moreover, the experiments only one GFN variant --- the detailed-balance one, which is also used for OC-GFN. It would help validate the generality of OC if we had experiments showing it worked on these different settings. Moreover, I'd be curious to know how other pretrained amortized sampling baselines (eg, VAEs, normalizing flows) fare against OC-GFN ---\xa0and what about pretraining a GFN on task A (without OC) and fine-tuning it on task B?\
- (minor) The second and fourth paragraphs of Section 4.2 mention the "reasoning potential" of GFNs, and that intractable marginalization leads to "slow thinking". Are these anthropomorphisms really needed for this paper?\
- (minor) I wished the preliminaries (Section 2) included a training objective like Eq. 5 & 9, and that these more clearly specified which are the optimization variables.\
- Some typos, there maybe more:\
- p. 3: multi-objective what?\
- p. 4: "given a reward R a posterior as a function"\
- p. 4: autotelicly -> autotelically?\
- p. 5: "in log-scale obtained from Eq. (5)" should be Eq. 4?'


The exposition is generally clear, and I enjoyed reading the paper. The authors first present the goal-conditioning idea and how it applies to GFNs, then walk the reader through their derivation and assumptions for amortized adaptation. I especially appreciated Section 2 which gave a clear and concise background.\
- The paper tackles an impactful problem for GFNs. While the pretraining solution is not particularly novel, it's a neat application of goal-condition RL to an amortized sampling problem. The authors also figured out how to make it work on a wide range of problems, and provide several ablations in the main text and the appendix.\
- The insight that a new sampling policy can be readily obtained from an outcome-conditioned flow is neat and, as far as I can tell, novel. This could spawn interest in outcome-conditioned flows and different ways to amortize Eq. 6.




Output:
{
Questions: [
{
\"Paper_id\": Asdho34,
\"review_id\": ioedh45,
\"Q_Number\": 1,
\"Question\":  "Please comment on the weaknesses outlined above.\
- Figures 10 and 11, right: Why is adaptation slower for OC-GFN than GFN in the first few thousand iterations? This is surprising since one would hope pretraining helps bootstrap downstream performance as in vision / language / RL. If it's an exploration phase, did you validate it and is there a way to side-step it?"
"
},


{
\"Paper_id\": Asdho34,
\"review_id\": ioedh45,
\"Q_Number\": 2,
\"Question\":  "There should be a discussions of assumptions behind the OC-GFNs pretraining. Namely, that transfer is only possible when the reward function changes but not if the action-space or the state-space change"
},
{
\"Paper_id\": Asdho34,
\"review_id\": ioedh45,
\"Q_Number\": 3,
\"Question\":  "These limitations (together with the applicability mentioned at the end of A.2) could be stated explicitly in the main text, and left to future work."
},


{
\"Paper_id\": Asdho34,
\"review_id\": ioedh45,
\"Q_Number\": 4,
\"Question\":  "While there are enough benchmarks, I believe none include continuous action/state spaces. Moreover, the experiments only one GFN variant --- the detailed-balance one, which is also used for OC-GFN. It would help validate the generality of OC if we had experiments showing it worked on these different settings"
},






{
\"Paper_id\": Asdho34,
\"review_id\": ioedh45,
\"Q_Number\": 5,
\"Question\":  "I'd be curious to know how other pretrained amortized sampling baselines (eg, VAEs, normalizing flows) fare against OC-GFN ---\xa0and what about pretraining a GFN on task A (without OC) and fine-tuning it on task B?"
},


{
\"Paper_id\": Asdho34,
\"review_id\": ioedh45,
\"Q_Number\": 6,
\"Question\":  "The second and fourth paragraphs of Section 4.2 mention the "reasoning potential" of GFNs, and that intractable marginalization leads to "slow thinking". Are these anthropomorphisms really needed for this paper?"
},
{
\"Paper_id\": Asdho34,
\"review_id\": ioedh45,
\"Q_Number\": 7,
\"Question\":  "I wished the preliminaries (Section 2) included a training objective like Eq. 5 & 9, and that these more clearly specified which are the optimization variables"},


{
\"Paper_id\": Asdho34,
\"review_id\": ioedh45,
\"Q_Number\": 8,
\"Question\":  "Some typos, there maybe more:\
- p. 3: multi-objective what?\
- p. 4: "given a reward R a posterior as a function"\
- p. 4: autotelicly -> autotelically?\
- p. 5: "in log-scale obtained from Eq. (5)" should be Eq. 4?"
},




]


}"""
\end{lstlisting}
\end{document}